\documentclass[nohyperref]{article}
\usepackage[final,main]{neurips_2026}
\usepackage{natbib}
\setcitestyle{numbers}
\setcitestyle{square}
\setcitestyle{comma}
\usepackage{multirow}
\usepackage[utf8]{inputenc}
\usepackage[T1]{fontenc}
\usepackage{url}
\usepackage{booktabs}
\usepackage{amsfonts}
\usepackage{microtype}
\usepackage{xcolor}
\usepackage{algorithm}
\usepackage{algorithmic}
\usepackage{makecell}
\usepackage{amsthm}
\usepackage{amssymb}
\usepackage{amsmath}
\usepackage{graphicx}
\theoremstyle{plain}
\usepackage{ulem}
\usepackage{colortbl,color}
\usepackage[hidelinks]{hyperref}
\definecolor{citeColor}{RGB}{0,20,115}
\hypersetup{colorlinks,linkcolor={citeColor},citecolor={citeColor},urlcolor={citeColor}}  
\usepackage{mathtools}
\usepackage{pifont}
\usepackage{tikz}

\usepackage{array}
\usepackage{enumerate}
\usepackage{tabularx}
\usepackage{enumitem}
\usepackage{caption}
\usepackage[toc,page,header]{appendix}
\usepackage[most]{tcolorbox}

\renewcommand{\emph}[1]{\textit{#1}}
\theoremstyle{plain}
\newtheorem{theorem}{Theorem}[section]
\newtheorem{lemma}[theorem]{Lemma}
\newtheorem{proposition}[theorem]{Proposition}

\usepackage{etoc}
\etocdepthtag.toc{mtchapter}
\etocsettagdepth{mtchapter}{subsection}
\etocsettagdepth{mtappendix}{none}

\title{
When Credit Turns Against Tools: Transition-Aware Credit Assignment in Agentic Reinforcement Learning
}

\makeatletter
\renewcommand*{\@fnsymbol}[1]{\ensuremath{\ifcase#1\or \dagger\or \ddagger\or
		\mathsection\or \mathparagraph\or \|\or **\or \dagger\dagger
		\or \ddagger\ddagger \else\@ctrerr\fi}}
\makeatother

\author{
	\textbf{Xiangyu Lu}$^{1}$ \quad
    \textbf{Zhanke Zhou}$^{1}$ \quad
    \textbf{Jiazhe Ning}$^{1,2}$ \quad 
	\textbf{Chentao Cao}$^{1}$ \quad\\
    \textbf{Jiangchao Yao}$^{3}$ \quad
	\textbf{Bo Han}$^{1}\thanks{Correspondence to Bo Han (bhanml@comp.hkbu.edu.hk).}$\
	\vspace{1mm} \\
	$^{1}$TMLR Group, Department of Computer Science, Hong Kong Baptist University \\
    $^{2}$Peking University 
	$^{3}$Cooperative Medianet Innovation Center, Shanghai Jiao Tong University 
}

\begin{document}

\maketitle

\begin{abstract}
Agentic reinforcement learning with verifiable rewards trains language models through multi-turn tool interactions, yet their reasoning performance and tool-calling ability can decline during training. We trace this degradation to three coupled mechanisms: (1) \textit{tool-token imbalance} produces negative overall tool credit when incorrect trajectories contain more tool tokens per turn than correct ones; (2) \textit{aggregation amplification} can further strengthen this negative signal as token-mean aggregation gives longer incorrect trajectories greater weight in policy updates; and (3) \textit{failure-driven retries} prolong trajectories as models tend to call tools again after execution errors, extending potential negative credit to recovery attempts. To counter this suppression, we propose \textbf{Transition-Aware Credit Assignment (TACA)}, which targets the transition from reasoning to tool use through supervision of both call initiation and generation. Specifically, TACA uses \textit{Advantage Projection} to enforce a nonnegative mean advantage on tool-call tokens within each prompt group through a minimal uniform shift. Additionally, TACA guides tool-use decisions with \textit{Decision Contrast}, comparing rewards for tool-assisted and direct answers to the same prompt. Through \textit{Supervised Replay}, it reinforces tool-call generation using calls from previously sampled correct trajectories. Across models and tasks, TACA improves performance and sustains more reliable tool interaction during training, raising mean accuracy over MT-GRPO on six reasoning benchmarks by 4.61 percentage points for Qwen2.5-1.5B and 3.61 for Qwen2.5-3B. Our code is available at \url{https://github.com/tmlr-group/TACA}.
\end{abstract}

\section{Introduction}
\label{sec.intro}

Reinforcement learning with verifiable rewards (RLVR) has improved reasoning in language models by using final-answer correctness as a terminal reward~\citep{cobbe2021training,uesato2022processoutcome,shao2024deepseekmath,guo2025deepseekr1,kimi2025k15,zeng2025simplerl,yu2025dapo,hu2025openreasonerzero,xie2025logicrl,wen2025rlvrreasoning}. Agentic RLVR extends this approach to settings with multi-turn tool interactions, where models use the observations returned by tools to guide subsequent reasoning~\citep{xue2025simpletir,jin2025searchr1,wang2025ragen,feng2025retool,qian2025toolrl,zhang2025tooln1,shang2025rstar2,mai2025artist}. Multi-turn Group Relative Policy Optimization (MT-GRPO)~\citep{shao2024deepseekmath,feng2025gigpo,luo2025agentlightning}, a widely used approach to agentic RLVR, assigns each trajectory's terminal reward to every model-generated turn. Within a trajectory, the tokens that invoke a tool and those that reason from its output therefore share the same advantage.

\begin{figure}[!t]
\centering
\includegraphics[width=\textwidth]{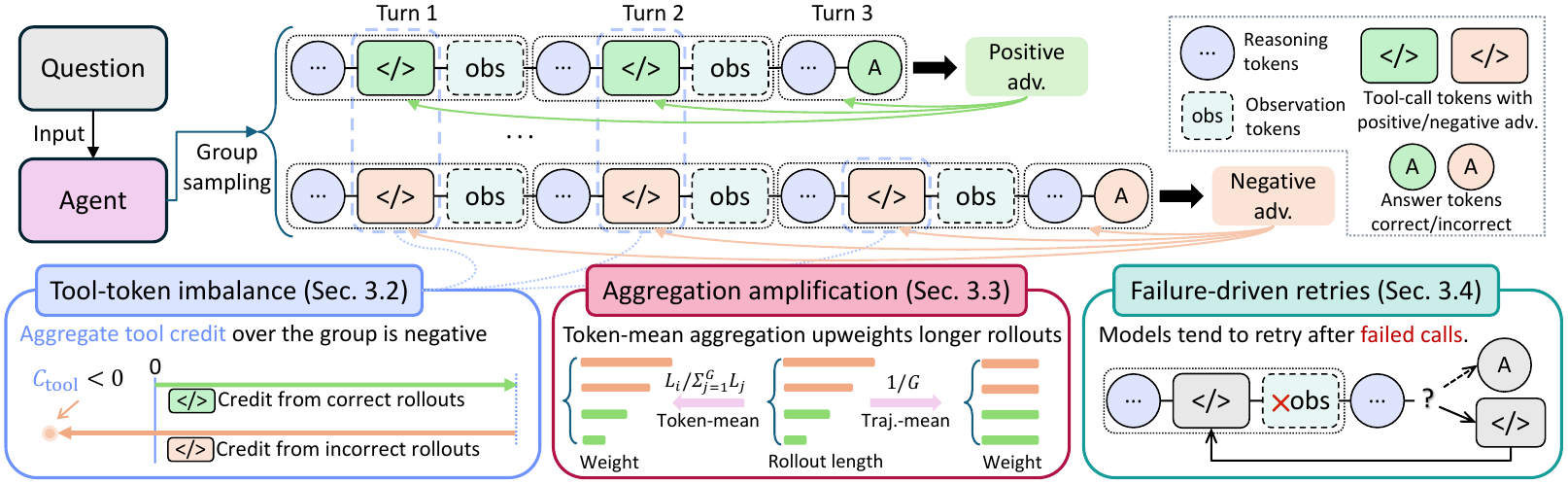}
\caption{Coupled mechanisms of training degradation. Terminal supervision can yield negative aggregate tool credit. Token-mean aggregation can further amplify this signal by upweighting longer incorrect rollouts, while retries prolong these rollouts and extend negative credit to recovery attempts.}
\label{fig:motivation}
\end{figure}

Despite these advances, training instability has been repeatedly documented in agentic reinforcement learning systems~\citep{xue2025simpletir,wang2025ragen,wang2025practitioner,dong2025aepo,xu2025epo,gu2025dspo,yu2025demystifying,deng2025lld}. Specifically, reasoning accuracy can improve early in training and then fall below the non-agentic baseline. During this decline, tool-use rates can remain broadly stable even as a smaller fraction of calls execute successfully. The model observes whether a call succeeds within the current turn, but MT-GRPO assigns credit to that call based on the final answer, potentially several turns later. Under this credit assignment, the same successful call can be reinforced in a trajectory that ends correctly and discouraged in one that ends incorrectly. This raises a natural question: \emph{how do the positive and negative tool-call credits combine across trajectories, and can they explain why tool execution and reasoning performance deteriorate with continued training?}

We therefore analyze how positive and negative advantages combine across the tokens that form tool calls in MT-GRPO (Figure~\ref{fig:motivation}). We show that when incorrect trajectories contain more tool-call tokens per generated turn than correct ones, this \emph{tool-token imbalance} produces negative overall credit for tool calls, even when correct trajectories constitute a majority. Additionally, \emph{aggregation amplification} can strengthen this negative signal, as token-mean aggregation gives negative credit from longer incorrect trajectories greater weight in policy updates than it would receive under equal trajectory weighting. Finally, models tend to retry on the next turn after a tool call fails. These \emph{failure-driven retries} occur more frequently in incorrect trajectories, where each additional tool-calling turn extends negative credit to the recovery attempt and gives the trajectory greater weight in the policy update.

Motivated by these findings, we propose \textbf{T}ransition-\textbf{A}ware \textbf{C}redit \textbf{A}ssignment (\textbf{TACA}), which augments MT-GRPO with targeted supervision for the transition from reasoning to tool use. TACA trains both the decision to seek external feedback and the generation of the call that determines what the model observes next. Specifically, \emph{Advantage Projection} offsets negative mean advantage on tool-call tokens while preserving reward-dependent advantage differences within each prompt group. \emph{Decision Contrast} guides tool-use decisions by comparing rewards between tool-using and direct-answer trajectories sampled for the same prompt. \emph{Supervised Replay} reuses previously sampled tool calls as supervision targets across policy updates. Together, these components train the transition from reasoning to tool interaction using evidence beyond the final outcome of an individual trajectory.

We evaluate TACA on four Qwen and Llama models across mathematical reasoning, coding, and decision-making. TACA achieves the best average accuracy in the Qwen2.5-1.5B and 3B baseline comparisons and improves over turn-level GRPO on Qwen2.5-7B and Llama-3.2-1B. With Qwen2.5-1.5B, it raises the TACO unit-test score from 41.98 to 53.70 and WebShop success rate from 56.8\% to 67.2\%. These gains accompany more reliable tool execution during training, with ablations supporting the complementary roles of advantage adjustments and replay supervision.

The main contributions of this work are:
\begin{itemize}[leftmargin=*]
\vspace{-6pt}
\item We identify negative aggregate tool credit as a mechanism underlying training degradation in multi-turn GRPO. Our analysis links its emergence to tool-token imbalance, with length weighting amplifying its influence and failure-driven retries extending negative supervision (Sec.~\ref{sec.analysis}).
\vspace{-3pt}
\item We propose TACA, a transition-aware objective that jointly optimizes tool-use decisions and tool-call generation by correcting negative aggregate tool credit, guiding tool-use decisions through reward contrasts, and maintaining tool-use capabilities through replay supervision (Sec.~\ref{sec.method}).
\vspace{-3pt}
\item We demonstrate that TACA consistently improves performance over existing RL methods across model families and tasks spanning mathematical reasoning, coding, and decision-making. TACA also sustains reasoning performance and tool-execution reliability as training progresses (Sec.~\ref{sec.exp}).
\end{itemize}

\section{Preliminaries}
\label{sec.preli}
\vspace{-6pt}

\paragraph{Multi-turn rollouts and advantages.}
In multi-turn interaction, each model-generated turn may contain reasoning, a tool call, or an answer; returned observations enter the context for subsequent turns. For each prompt $x$, a fixed rollout policy $\pi_{\theta_{\mathrm{old}}}$ samples $G$ trajectories $\{\tau_i\}_{i=1}^{G}$ where trajectory $\tau_i$ contains $S_i$ generated turns and receives a bounded scalar terminal reward $R_i$.

Group Relative Policy Optimization (GRPO)~\citep{shao2024deepseekmath} computes advantages by standardizing rewards within each prompt's sample group.
We study multi-turn GRPO (MT-GRPO) with turn-level reward normalization~\citep{luo2025agentlightning,feng2025gigpo,zhang2025rlvmr,qin2025spear,jiang2025lamer,wang2026envrl}: each trajectory reward $R_i$ is replicated across its $S_i$ turns before normalization. With $M=\sum_i S_i$ total turns, the reward mean and variance are
\begin{equation}
\mu_{\mathrm{turn}}=\frac{1}{M}\sum_i S_iR_i,
\qquad
\sigma_{\mathrm{turn}}^2=\frac{1}{M}\sum_i S_i(R_i-\mu_{\mathrm{turn}})^2.
\label{eq:turn_reward_statistics}
\end{equation}

\vspace{-6pt}
Every generated token in $\tau_i$ then receives the same advantage $A_i^{\mathrm{turn}}=(R_i-\mu_{\mathrm{turn}})/\sigma_{\mathrm{turn}}$.

\vspace{-6pt}
\paragraph{Token-mean policy objective.}
Let $\ell_{i,h}$ denote the number of generated tokens scored by the objective in turn $h$ of trajectory $i$, and $N=\sum_i\sum_{h=1}^{S_i}\ell_{i,h}$ their total count. Observations provide context but are excluded from the objective.
We use token-mean aggregation~\citep{yu2025dapo,li2025torl,jin2025searchr1,xue2025simpletir,xi2025agentgymrl}. The current policy $\pi_\theta$ maximizes the objective averaged over these $N$ tokens:
\begin{equation}
\mathcal{J}_{\mathrm{MT\text{-}GRPO}}(\pi_\theta)
\triangleq
\frac{1}{N}\sum_{i=1}^{G}\sum_{h=1}^{S_i}\sum_{t=1}^{\ell_{i,h}}
\Big[\phi(\rho_{i,h,t},A_i^{\mathrm{turn}})-\beta\,\mathbb{D}_{\mathrm{KL}}\!\left[\pi_\theta\,\big\|\,\pi_{\mathrm{ref}}\right]_{i,h,t}\Big],
\label{eq:turn_tok_objective}
\end{equation}
Here, $\rho_{i,h,t}$ is the token-level importance sampling ratio of the current policy relative to the rollout policy. The clipped surrogate $\phi(\rho,a)=\min\{\rho a,\operatorname{clip}(\rho,1-\varepsilon,1+\varepsilon)a\}$ uses advantage $a$ and clipping threshold $\varepsilon$. The KL penalty has weight $\beta$ and fixed reference policy $\pi_{\mathrm{ref}}$. 

\section{Understanding Training Degradation in Multi-Turn GRPO}
\label{sec.analysis}

\vspace{-6pt}

We analyze how terminal-reward supervision, reward normalization, and objective aggregation determine credit on tool interactions. We first derive the scalar credit and trajectory-weighting relationships, then examine how retry behavior contributes to these effects. Details are in Appendix~\ref{app:extended_analysis}.

\vspace{-6pt}

\subsection{Training Degradation and Tool Interaction}
\label{ssec:collapse_observation}
\vspace{-6pt}

\begin{figure*}[t!]
\centering
\includegraphics[width=\textwidth]{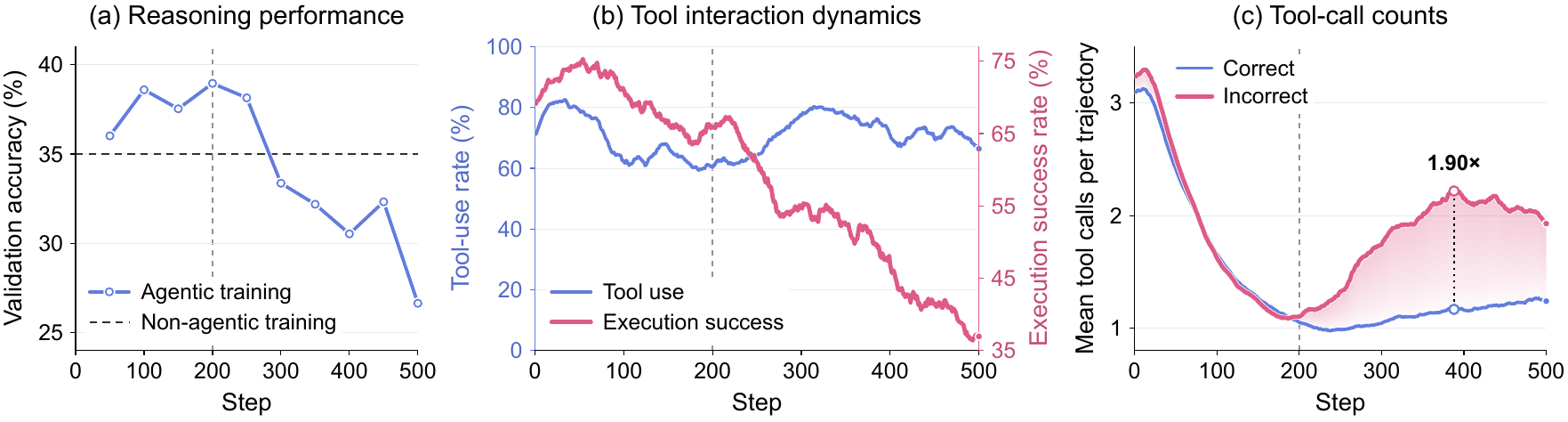}
\vspace{-18pt}
\caption{Training dynamics of MT-GRPO.
(a) Validation accuracy peaks and then falls below the non-agentic reference.
(b) Tool-execution success declines despite broadly stable tool-use rates.
(c) Mean tool-call counts rise for incorrect trajectories and remain relatively stable for correct ones.}
\vspace{-8pt}
\label{fig:sec3_overview}
\end{figure*}

Training instability in multi-turn agentic RL has been documented across recent systems~\citep{xue2025simpletir,jin2025searchr1,wang2025ragen,wang2025practitioner,dong2025aepo,xu2025epo,gu2025dspo,yu2025demystifying,deng2025lld,hao2026toolcollapse}. We reproduce this degradation in a widely adopted tool-integrated reasoning setting, training Qwen2.5-3B-Instruct on DAPO-Math with MT-GRPO. After an initial improvement, validation accuracy declines and falls below the non-agentic reference (Figure~\ref{fig:sec3_overview}a). Meanwhile, tool-execution success declines while tool-use rates remain broadly stable (Figure~\ref{fig:sec3_overview}b), and incorrect trajectories contain more tool calls on average than correct ones (Figure~\ref{fig:sec3_overview}c). These observations suggest a close connection between training degradation and the quality of tool interaction.

\vspace{-6pt}
\subsection{Terminal-Reward Supervision and Negative Tool Credit}
\label{ssec:token_mean_bias}
\vspace{-6pt}

MT-GRPO assigns each trajectory's terminal-reward advantage to all its scored tokens. Turn-level normalization enforces $\sum_i S_i A_i^{\mathrm{turn}}=0$, balancing positive and negative advantage across generated turns. This balance can break on tool tokens: when incorrect trajectories contain more tool tokens per turn than correct ones, negative coefficients dominate the aggregate tool credit.

Consider a prompt group containing both correct ($R_i=1$) and incorrect ($R_i=0$) trajectories. Let $m_i$ count the model-generated tokens forming tool calls in trajectory $i$. Let $\bar m_{\mathrm{turn}}^{+}$ and $\bar m_{\mathrm{turn}}^{-}$ denote the average number of tool tokens per generated turn in correct and incorrect trajectories, respectively.

\begin{proposition}[Negative tool-credit condition]
\label{prop:tool_credit}
In the fully on-policy setting, the reward surrogate assigns coefficient $A_i^{\mathrm{turn}}/N$ to each tool token's log-probability gradient in trajectory $i$. Summing these coefficients gives
\begin{equation}
C_{\mathrm{tool}}
\coloneqq\frac{1}{N}\sum_i m_iA_i^{\mathrm{turn}}
=\frac{M\sigma_{\mathrm{turn}}}{N}
\left(\bar m_{\mathrm{turn}}^{+}-\bar m_{\mathrm{turn}}^{-}\right),
\label{eq:tool_credit}
\end{equation}
where $M$ denotes the total number of generated turns and $N$ the total number of tokens included in the policy objective. Consequently, $C_{\mathrm{tool}}<0$ if and only if incorrect trajectories contain more tool tokens per generated turn than correct trajectories.
\end{proposition}

\begin{figure*}[t]
\centering
\includegraphics[width=\textwidth]{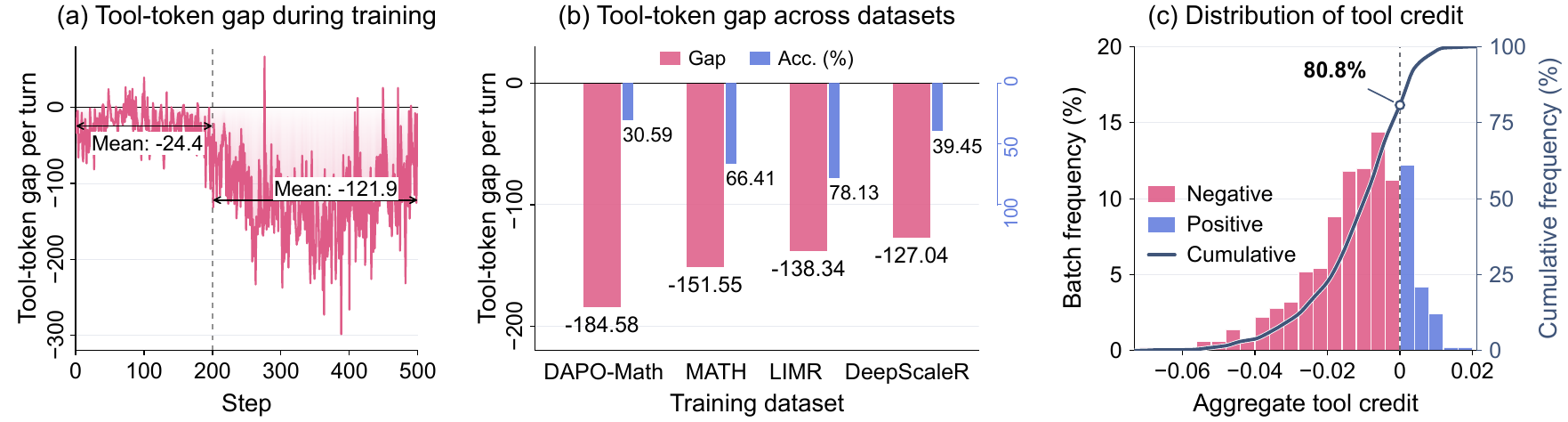}
\vspace{-18pt}
\caption{Negative credit on tool tokens.
(a) The tool-token gap $\bar m_{\mathrm{turn}}^{+}-\bar m_{\mathrm{turn}}^{-}$ in tool tokens per turn becomes more negative during degradation.
(b) At step 300, this gap remains negative across training datasets with different accuracies.
(c) Aggregate tool credit $C_{\mathrm{tool}}$ is negative in most batches.}
\label{fig:sec3_mechanism}
\vspace{-12pt}
\end{figure*}

Proposition~\ref{prop:tool_credit} shows that negative tool credit can arise even when correct trajectories constitute a majority. Within a mixed-outcome group, the difference in tool tokens per turn determines the sign of aggregate credit, while reward composition also affects its magnitude. Figure~\ref{fig:sec3_mechanism}a,b characterize this imbalance during training and across datasets with different accuracies. Within such groups, terminal-reward supervision assigns the same negative advantage to every call in an incorrect trajectory, including calls that execute successfully before an error in subsequent reasoning. Tool-call credit is therefore tied to failures that may occur elsewhere in the trajectory. Under the condition in Proposition~\ref{prop:tool_credit}, these negative contributions collectively outweigh positive credit on tool tokens. Token-mean aggregation weights prompt-group credits by their total generated-token counts, and Figure~\ref{fig:sec3_mechanism}c measures the resulting batch credit using advantages computed within each prompt group.

\subsection{Length-Proportional Reweighting}
\label{ssec:acc_bias}
Beyond the tool-token imbalance, aggregate tool credit also depends on trajectory weighting. We analyze the shift induced by token-mean aggregation relative to equal trajectory weighting.

\vspace{-10pt}
\paragraph{Tool-credit aggregation.}
Let $\mathcal B$ index the trajectories in a training batch. For each $i\in\mathcal B$, let $L_i=\sum_{h=1}^{S_i}\ell_{i,h}>0$ count the model-generated tokens and $m_i$ count the tool-call tokens among them. With turn-level advantage $A_i^{\mathrm{turn}}$ computed within its prompt group, the trajectory contributes tool credit $(m_i/L_i)A_i^{\mathrm{turn}}$ per generated token. Token-mean aggregation weights this contribution by trajectory length, whereas trajectory-mean aggregation weights trajectories equally:
\begin{equation}
C_{\mathrm{tool}}^{\mathrm{token}}
=\sum_{i\in\mathcal B}
\frac{L_i}{\sum_{j\in\mathcal B}L_j}
\frac{m_i}{L_i}A_i^{\mathrm{turn}},
\qquad
C_{\mathrm{tool}}^{\mathrm{traj}}
=\frac{1}{|\mathcal B|}
\sum_{i\in\mathcal B}
\frac{m_i}{L_i}A_i^{\mathrm{turn}}.
\label{eq:tool_credit_aggregation}
\end{equation}

\begin{figure*}[!t]
\centering
\includegraphics[width=\textwidth]{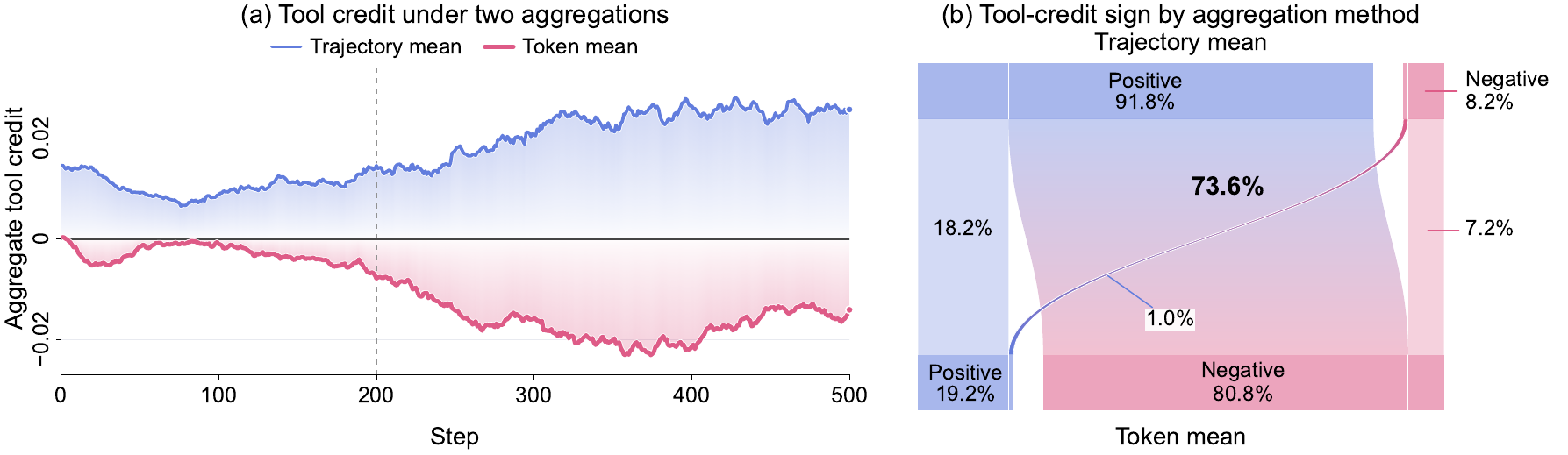}
\vspace{-18pt}
\caption{Length weighting and aggregate tool credit.
(a) Token-mean aggregation shifts credit downward relative to trajectory-mean aggregation.
(b) Length weighting turns positive tool credit negative in 73.6\% of batches. This shifts the balance toward penalizing sampled tool calls.}
\vspace{-8pt}
\label{fig:sec3_length_amplification}
\end{figure*}

\begin{figure*}[!t]
\centering
\includegraphics[width=\textwidth]{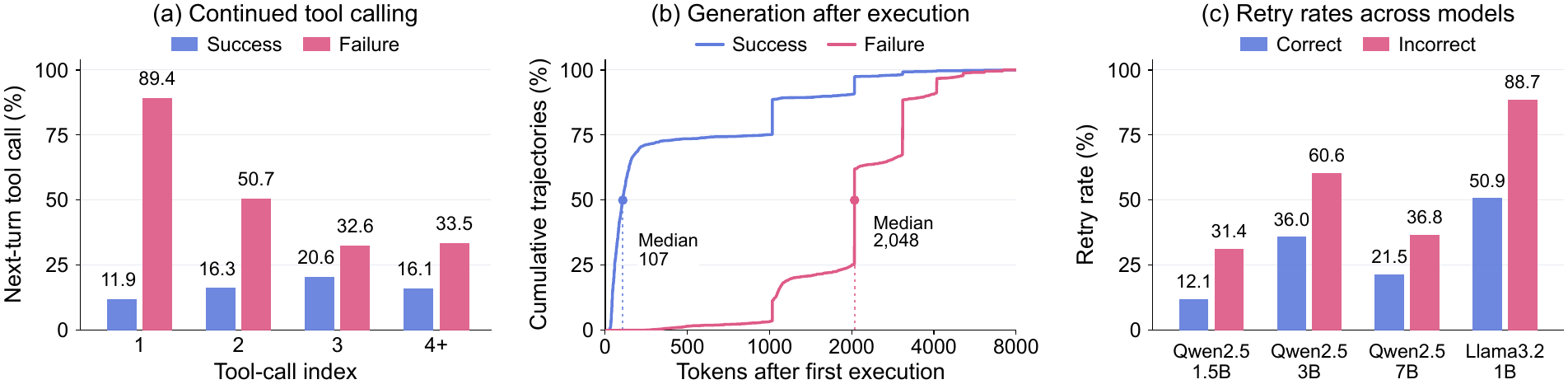}
\vspace{-18pt}
\caption{Retry behavior and continuation length.
(a) A tool call in the next turn is more likely after failed than successful executions.
(b) First-execution failure is associated with longer subsequent generation.
(c) Incorrect trajectories have higher retry rates across all four models on MATH.}
\label{fig:sec3_retry_feedback}
\vspace{-12pt}
\end{figure*}

\vspace{-10pt}
\paragraph{Aggregation-induced credit shift.}
Their difference isolates the effect of length weighting:
\begin{equation}
C_{\mathrm{tool}}^{\mathrm{token}}
-
C_{\mathrm{tool}}^{\mathrm{traj}}
=
\frac{
\operatorname{Cov}_{\mathcal B}
\!\left(L_i,\frac{m_i}{L_i}A_i^{\mathrm{turn}}\right)
}{
\mathbb E_{\mathcal B}[L_i]
},
\label{eq:tool_aggregation_covariance}
\end{equation}
where the empirical mean and covariance assign weight $1/|\mathcal B|$ to each trajectory. The covariance determines the direction of the credit shift: when longer trajectories tend to carry lower tool credit per generated token, token-mean aggregation shifts aggregate credit downward. Such a shift amplifies negative credit and reverses positive credit when its magnitude exceeds $C_{\mathrm{tool}}^{\mathrm{traj}}$.

Figure~\ref{fig:sec3_length_amplification}a shows the downward credit shift induced by length weighting, while Figure~\ref{fig:sec3_length_amplification}b identifies batches whose tool credit changes from positive to negative. In these batches, negative tool-token coefficients outweigh positive coefficients in the reward surrogate, shifting the balance toward updates that penalize sampled tool calls. Length weighting thus compounds the suppression analyzed in Section~\ref{ssec:token_mean_bias}: it increases the relative influence of negative supervision on tool interactions, including successful executions within ultimately incorrect trajectories.

\subsection{Failure-Driven Retries}
\label{ssec:closed_loop}

\paragraph{Failure-conditioned continuation.}
The trajectories underlying these credit imbalances are shaped by the model's response to execution errors. A retry adds another tool action and any accompanying reasoning before the terminal reward is assigned. Execution outcomes therefore influence which subsequent actions enter the training sample and how long generation continues. The higher next-turn tool-call rate after failed executions (Figure~\ref{fig:sec3_retry_feedback}a) and longer continuation following an initial execution failure (Figure~\ref{fig:sec3_retry_feedback}b) support this dependence of trajectory composition on intermediate outcomes.

\vspace{-10pt}
\paragraph{Terminal supervision of recovery attempts.}
Each decision to retry expands the set of actions sharing the completed trajectory's terminal-reward advantage. Within a mixed-outcome prompt group, an ultimately incorrect trajectory assigns negative advantage to both the failed call and subsequent recovery attempts, including retries that execute successfully. Figure~\ref{fig:sec3_retry_feedback}c shows higher retry rates in incorrect trajectories across models, connecting recovery behavior to the terminal outcomes that determine credit. The recovery behavior thus shapes subsequent supervision: in an unsuccessful trajectory, each additional recovery attempt adds tool actions receiving the same negative advantage.

\section{Transition-Aware Credit Assignment}
\label{sec.method}

\begin{figure}[!t]
\centering
\includegraphics[width=\textwidth]{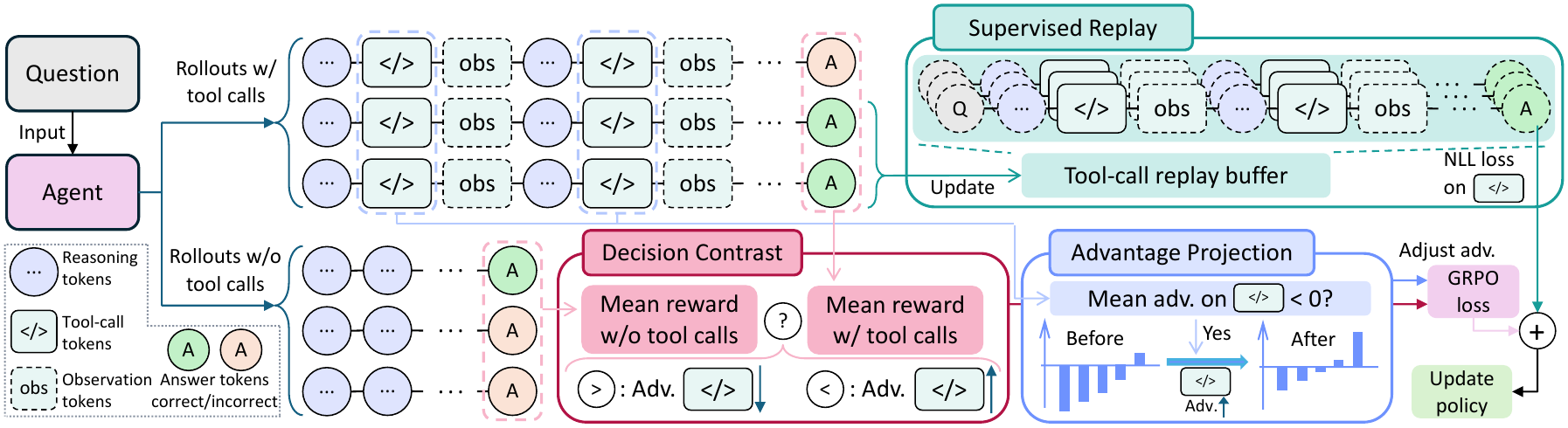}
\caption{Overview of TACA. TACA adjusts tool-call and decision-token advantages using within-prompt reward statistics from tool-using and direct-answer rollouts. These adjustments and replay supervision from correct trajectories jointly guide tool-use decisions and tool-call generation.}
\label{fig:sec4_method_overview}
\vspace{-12pt}
\end{figure}

Motivated by Section~\ref{sec.analysis}, we propose {Transition-Aware Credit Assignment (TACA)}, which augments MT-GRPO with token-level advantage corrections and auxiliary supervision for tool-call generation (Figure~\ref{fig:sec4_method_overview}). Concretely, TACA offsets negative mean advantages on tool-call tokens (Advantage Projection), guides tool-use decisions by comparing rewards between tool-using and direct-answer trajectories sampled for the same prompt (Decision Contrast), and reinforces tool-call generation through an auxiliary likelihood objective on previously sampled calls (Supervised Replay).

\subsection{Advantage Projection}

We target the negative aggregate credit identified in Section~\ref{sec.analysis} by applying a uniform advantage correction to tool-related tokens. For each prompt $x$, let $\mathcal Z$ denote the model-generated tokens forming tool calls across its sampled trajectories, and let $A_i$ denote the GRPO advantage of trajectory $i$. We compute the mean advantage over these tokens and the corresponding correction:
\begin{equation}
\bar A_{x,\mathcal Z}
=
\frac{1}{|\mathcal Z|}
\sum_{(i,h,t)\in\mathcal Z} A_i,
\qquad
b_{x,\mathcal Z}
=
\max\{-\bar A_{x,\mathcal Z},0\},
\label{eq:zmean_advantage}
\end{equation}
where $(i,h,t)$ indexes the trajectory, turn, and token position, and $|\mathcal Z|$ is the number of tool-call tokens in the set. Advantage Projection replaces $A_i$ with $A_i+b_{x,\mathcal Z}$ exclusively at positions in $\mathcal Z$.

The correction changes the aggregate advantage over $\mathcal Z$ from $|\mathcal Z|\bar A_{x,\mathcal Z}$ to $|\mathcal Z|\max\{\bar A_{x,\mathcal Z},0\}$. Negative aggregate credit is therefore neutralized, while positive aggregate credit is preserved. The resulting vector is the Euclidean projection of the original tool-call advantages onto the halfspace of vectors with nonnegative mean. Within the same group, the uniform offset preserves the advantage difference $A_i-A_j$ between tool tokens from trajectories $i$ and $j$. Advantage Projection therefore enforces nonnegative aggregate tool credit while retaining these reward-dependent differences.

\subsection{Decision Contrast}

We complement the tool-credit correction with a reward comparison that guides the choice between tool use and direct answering. For each prompt $x$, let $\mathcal T_x$ contain sampled trajectories that invoke at least one tool and $\mathcal N_x$ contain those that answer without tools. Let $\bar R_{\mathcal T_x}$ and $\bar R_{\mathcal N_x}$ denote their mean terminal rewards, counting each trajectory once. We define the reward contrast as
\begin{equation}
U_x
=
\frac{n_x}{n_x+\eta}
\left(\bar R_{\mathcal T_x}-\bar R_{\mathcal N_x}\right),
\qquad
n_x=\min\{|\mathcal T_x|,|\mathcal N_x|\}.
\label{eq:gic}
\end{equation}
The factor $n_x/(n_x+\eta)$ shrinks the reward contrast according to the smaller group size, limiting the correction when either reward estimate is based on few trajectories. The parameter $\eta\geq0$ controls the shrinkage strength; $U_x=0$ when either group is empty. We apply this contrast to tokens that specify the response action, concentrating the correction at the tool-calling  choice. Within each prompt group, let $\mathcal Z_{\mathrm{dec}}$ contain the tokens forming tool-call opening markers (e.g., \texttt{<python\_code>}) in $\mathcal T_x$, and $\mathcal Z_{\mathrm{dir}}$ contain the tokens forming answer-opening markers (e.g., \texttt{<answer>}) in $\mathcal N_x$. Decision Contrast assigns corrections $+U_x$ and $-U_x$ to token advantages in $\mathcal Z_{\mathrm{dec}}$ and $\mathcal Z_{\mathrm{dir}}$, respectively.

This correction changes the difference between tool and direct-answer decision tokens' mean advantages by $2U_x$. Higher sampled rewards for tool use therefore increase the advantage assigned to tool-decision tokens relative to direct-answer decision tokens; higher rewards for direct answering reverse the correction. The comparison thus provides a prompt-specific preference between the two response modes, with each trajectory contributing equally to its mode's reward estimate.

\subsection{Supervised Replay}

To retain tool-call generation during training, Supervised Replay reuses previously sampled calls as direct supervision targets. At each policy update, we add context-response pairs from tool-calling turns in trajectories with correct final answers to a fixed-capacity replay buffer, discarding the oldest entries when the capacity is exceeded. Each stored pair $(c,y)$ consists of the preceding interaction context $c$, including any tool observations, and the model-generated response $y$ containing a tool call.

We uniformly sample a replay minibatch $\mathcal D_{\mathrm{rep}}$ from this buffer. For each response $y$, a binary mask $m_t^{\mathrm{rep}}(c,y)$ identifies the tool-call tokens used as prediction targets. To train the current policy $\pi_\theta$ to reproduce these calls under their original contexts, we minimize their average negative log-likelihood:
\begin{equation}
\mathcal L_{\mathrm{rep}}(\theta)
=
-\frac{1}{M_{\mathrm{rep}}}
\sum_{(c,y)\in\mathcal D_{\mathrm{rep}}}
\sum_{t=1}^{|y|}
m_t^{\mathrm{rep}}(c,y)
\log\pi_\theta(y_t\mid c,y_{<t}),
\label{eq:support_anchor}
\end{equation}
where $M_{\mathrm{rep}}$ counts the tool-call tokens in the minibatch, and $y_t$ and $y_{<t}$ denote the current token and preceding response tokens, respectively. Preceding reasoning remains in the conditioning context.

Minimizing this loss reinforces the recorded tool-call continuations under their original contexts. Replay retains these targets across policy updates, so successful trajectories continue to supervise tool-call generation as the rollout distribution evolves, complementing the advantage adjustments.

\subsection{Training Objective}

\paragraph{Transition-aware advantages.}
We add the projection offset to tool-call token advantages and the signed Decision Contrast signal to decision-token advantages, applying both corrections at tool-call openings. For token $(i,h,t)$ in trajectory $i$ sampled for prompt $x$, the original advantage $A_i$ becomes
\begin{equation}
A^{\mathrm{TACA}}_{i,h,t}
=
A_i
+\underbrace{b_{x,\mathcal Z}\,\mathbf 1\{(i,h,t)\in\mathcal Z\}}_{\text{Advantage Projection}}
+\underbrace{\lambda_{\mathrm{dc}}U_x
\left[\mathbf 1\{(i,h,t)\in\mathcal Z_{\mathrm{dec}}\}
-\mathbf 1\{(i,h,t)\in\mathcal Z_{\mathrm{dir}}\}\right]}_{\text{Decision Contrast}},
\label{eq:taca_advantage}
\end{equation}
where $\mathbf 1\{\cdot\}$ denotes an indicator and $\lambda_{\mathrm{dc}}\geq0$ controls the strength of Decision Contrast.

\vspace{-6pt}
\paragraph{Policy optimization.}
The TACA advantages enter the token-mean policy objective, together with the auxiliary replay loss. For a rollout batch, let $S_i$ denote the number of generated turns in trajectory $i$, $\ell_{i,h}$ the number of model-generated tokens in turn $h$, and $N=\sum_i\sum_{h=1}^{S_i}\ell_{i,h}$ their total count. With $\lambda_{\mathrm{rep}}\geq0$ weighting replay supervision, the policy maximizes
\begin{equation}
\mathcal J_{\mathrm{TACA}}(\pi_\theta)
=
\frac{1}{N}
\sum_i\sum_{h=1}^{S_i}\sum_{t=1}^{\ell_{i,h}}
\left[
\phi\!\left(\rho_{i,h,t},A^{\mathrm{TACA}}_{i,h,t}\right)
-\beta\,\mathbb D_{\mathrm{KL}}
\!\left[\pi_\theta\middle\|\pi_{\mathrm{ref}}\right]_{i,h,t}
\right]
-\lambda_{\mathrm{rep}}\mathcal L_{\mathrm{rep}}(\theta).
\label{eq:taca_objective}
\end{equation}

\vspace{-6pt}
Within each prompt group, TACA preserves reward-dependent advantage differences between tokens receiving the same adjustment, while allowing reasoning and tool interaction to receive different supervision. An unsuccessful trajectory can receive an upward adjustment at tool initiation when the reward contrast favors tool use, while tool calls from correct trajectories provide replay supervision across updates. The transition from reasoning to tool interaction is trained through response comparisons and replay from correct trajectories, using evidence beyond one rollout's outcome.

\providecolor{TACAtint}{HTML}{F0F3FD}
\section{Experiments}
\label{sec.exp}
\vspace{-8pt}

\begin{table*}[t]
\centering
\small
\renewcommand{\arraystretch}{1.0}
\setlength{\aboverulesep}{2pt}
\setlength{\belowrulesep}{2pt}
\setlength{\tabcolsep}{5pt}
\caption{Main results of post-training experiments (accuracy, in \%). Bold and underlined values indicate the best and second-best results within each model, respectively.}
\vspace{-8pt}
\label{tab:main_results_all}
\resizebox{\textwidth}{!}{
\begin{tabular}{lccccccc}
\toprule
\multirow{2}{*}{\makecell{\textbf{Methods}}}
& \multicolumn{6}{c}{\textbf{Benchmarks}}
& \multirow{2}{*}{\makecell{\textbf{Average}}} \\
\cmidrule(lr){2-7}
& \makecell{AIME24} & \makecell{AIME25} & \makecell{MATH500} & \makecell{AMC23} & \makecell{OlympiadBench} & \makecell{GSM8K}\\
\midrule
\multicolumn{8}{c}{\textbf{Qwen2.5-1.5B-Instruct}} \\
\cmidrule(lr){1-8}
\multicolumn{8}{l}{\textit{Training-free}} \\
Base model & 1.09 & 0.35 & 38.83 & 15.56 & 10.52 & 59.42 & 20.96 \\
Agentic prompting & 0.20 & 0.0 & 10.46 & 3.44 & 2.69 & 16.29 & 5.51 \\
\cmidrule(lr){1-8}
\multicolumn{8}{l}{\textit{Non-agentic training}} \\
GRPO & 3.03 & 0.69 & 50.10 & 26.02 & 17.60 & \underline{72.55} & 28.33 \\
\cmidrule(lr){1-8}
\multicolumn{8}{l}{\textit{Agentic training}} \\
ReTool & \underline{3.81} & 1.08 & \underline{51.90} & 27.34 & 16.65 & 70.74 & 28.59 \\
Turn-level GRPO & 3.26 & 0.97 & 50.70 & \textbf{27.50} & \underline{17.88} & 71.49 & \underline{28.63} \\
SimpleTIR & 3.04 & \underline{1.38} & 49.54 & 22.99 & 17.39 & 70.48 & 27.47 \\
\rowcolor{TACAtint}
\textbf{TACA (ours)} & \textbf{8.43} & \textbf{8.12} & \textbf{59.14} & \underline{27.42} & \textbf{22.20} & \textbf{74.15} & \textbf{33.24} \\
\midrule
\multicolumn{8}{c}{\textbf{Qwen2.5-3B-Instruct}} \\
\cmidrule(lr){1-8}
\multicolumn{8}{l}{\textit{Training-free}} \\
Base model & 3.45 & 1.04 & 56.65 & 29.32 & 22.15 & 76.76 & 31.56 \\
Agentic prompting & 4.16 & 1.13 & 38.52 & 17.43 & 14.48 & 45.27 & 20.17 \\
\cmidrule(lr){1-8}
\multicolumn{8}{l}{\textit{Non-agentic training}} \\
GRPO & 5.55 & 2.29 & \underline{62.45} & 41.02 & 24.99 & 81.94 & 36.37 \\
\cmidrule(lr){1-8}
\multicolumn{8}{l}{\textit{Agentic training}} \\
ReTool & 6.04 & 6.62 & 61.03 & 36.59 & 27.34 & 82.95 & 36.76 \\
Turn-level GRPO & \underline{13.57} & \underline{11.21} & 58.87 & \underline{47.67} & 28.51 & 79.57 & \underline{39.90} \\
SimpleTIR & 10.76 & 9.51 & 61.37 & 41.16 & \underline{31.67} & \textbf{83.52} & 39.67 \\
\rowcolor{TACAtint}
\textbf{TACA (ours)} & \textbf{15.56} & \textbf{15.00} & \textbf{64.51} & \textbf{47.96} & \textbf{34.62} & \underline{83.42} & \textbf{43.51} \\
\bottomrule
\end{tabular}
}
\vspace{-16pt}
\end{table*}

\begin{table}[t]
\centering
\caption{Coding and interactive task performance on Qwen2.5-1.5B-Instruct. TACO reports unit-test scores by source and their unweighted average. WebShop reports success rate (SR) and task score.}
\label{tab:task_results}\label{tab:taco}\label{tab:webshop}\label{tab:taco_sources}
\begingroup
\fontsize{9}{11}\selectfont
\setlength{\aboverulesep}{2pt}
\setlength{\belowrulesep}{2pt}
\renewcommand{\arraystretch}{1.0}
\setlength{\tabcolsep}{0pt}
\begin{tabular}{@{}p{72pt}>{\centering\arraybackslash}p{52pt}>{\centering\arraybackslash}p{54pt}>{\centering\arraybackslash}p{50pt}>{\centering\arraybackslash}p{44pt}>{\centering\arraybackslash}p{34pt}>{\centering\arraybackslash}p{45pt}>{\centering\arraybackslash}p{45pt}@{}}
\toprule
\multirow{2}{*}[-2.5pt]{\textbf{Method}} & \multicolumn{5}{c}{TACO} & \multicolumn{2}{c}{WebShop} \\
\cmidrule(lr){2-6}\cmidrule(l){7-8}
& HackerRank & HackerEarth & CodeForces & CodeChef & Avg. & SR (\%) & Score \\
\midrule
MT-GRPO & 59.80 & 25.00 & 48.13 & 34.98 & 41.98 & 56.80 & 75.80 \\
GiGPO & -- & -- & -- & -- & -- & 65.00 & 83.10 \\
\rowcolor{TACAtint}
\textbf{TACA} & \textbf{64.71} & \textbf{50.00} & \textbf{58.98} & \textbf{41.12} & \textbf{53.70} & \textbf{67.20} & \textbf{85.80} \\
\bottomrule
\end{tabular}
\endgroup
\vspace{-16pt}
\end{table}

\subsection{Experimental Setup}
\label{ssec:setup}
\vspace{-6pt}
\paragraph{Training setup.}
We post-train Qwen2.5-1.5B-Instruct, Qwen2.5-3B-Instruct, Qwen2.5-7B-Instruct~\citep{yang2024qwen2}, and Llama-3.2-1B-Instruct on MATH~\citep{hendrycksmath2021}. We use AlphaApollo~\citep{zhou2025alphaapollo} for multi-turn agentic training, with code execution as the primary reasoning tool. Training uses four A800 GPUs, 32 prompts per batch, eight rollouts per prompt, and a learning rate of $10^{-6}$. Trajectories contain at most six generated turns, with up to 2,048 response tokens per turn and a three-second execution timeout. We also train Qwen2.5-1.5B-Instruct separately on TACO~\citep{li2023taco} and WebShop~\citep{yao2022webshop} to evaluate coding and interactive decision-making, respectively. All tasks use terminal rewards.

\vspace{-10pt}
\paragraph{Evaluation setup.}
We evaluate on MATH500~\citep{hendrycksmath2021}, AIME 2024, AIME 2025, AMC 2023, OlympiadBench~\citep{he2024olympiadbench}, and GSM8K~\citep{cobbe2021training}. We sample 32 responses per question at temperature $0.6$ with top-p sampling and report benchmark accuracies and their unweighted mean. TACO reports unit-test scores on its test set, while WebShop reports success rate and task score on its official split.

\vspace{-10pt}
\paragraph{Baselines.}
We compare TACA with non-agentic GRPO~\citep{shao2024deepseekmath} and agentic methods ReTool~\citep{feng2025retool}, MT-GRPO~\citep{feng2025gigpo}, GiGPO~\citep{feng2025gigpo}, and SimpleTIR~\citep{xue2025simpletir}. Base-model inference and agentic prompting serve as training-free references. Training and evaluation settings are matched for each model-task pair. Further experimental details for TACA and the baseline methods are provided in Appendix~\ref{app:implementation}.

\vspace{-6pt}
\subsection{Main Results}
\label{ssec:model_generalization}
\vspace{-6pt}

\paragraph{TACA improves the effectiveness of agentic training.}
Table~\ref{tab:main_results_all} shows that TACA achieves the highest average accuracy on both Qwen2.5-1.5B-Instruct and Qwen2.5-3B-Instruct, reaching 33.24\% and 43.51\%, with the best score on ten of twelve model--benchmark pairs. Agentic prompting alone lowers average accuracy relative to the base models, whereas TACA exceeds non-agentic GRPO by 4.91 and 7.14 percentage points, respectively. Among the agentic training methods, TACA also improves over MT-GRPO by 4.61 and 3.61 percentage points. \emph{These gains over MT-GRPO suggest that TACA makes terminal-reward supervision more effective for learning tool-assisted reasoning.}

\vspace{-10pt}
\paragraph{Reasoning gains extend to challenging competition problems.}
On Qwen2.5-3B-Instruct, TACA exceeds the strongest competing result on AIME 2024, AIME 2025, and OlympiadBench by 1.99, 3.79, and 2.95 percentage points, respectively (Table~\ref{tab:main_results_all}). The relative gains are particularly pronounced for the 1.5B model, which improves on the strongest baseline from 3.81\% to 8.43\% on AIME 2024 and from 1.38\% to 8.12\% on AIME 2025. The resulting 1.5B TACA policy also outperforms the larger 3B ReTool model on both AIME benchmarks. \emph{This comparison suggests that better agentic training can help smaller models close the gap with larger models on difficult reasoning tasks.}

\vspace{-10pt}
\paragraph{TACA improves functional correctness across coding problem sources.}
\label{ssec:cross_domain}
Table~\ref{tab:task_results} compares coding policies on the TACO test set. TACA raises the mean unit-test score from 41.98 to 53.70, improving all four problem sources: HackerRank, HackerEarth, CodeForces, and CodeChef. The gains range from 4.91 points on HackerRank to 25.00 on HackerEarth. \emph{TACA's training objective remains effective when the training signal measures the functional correctness of generated programs.}

\vspace{-10pt}
\paragraph{TACA improves successful completion in interactive decision-making.}
On WebShop, TACA increases success rate from 56.8\% to 67.2\% and task score from 75.8 to 85.8 relative to MT-GRPO (Table~\ref{tab:task_results}). It also exceeds GiGPO by 2.2 percentage points in success rate and 2.7 points in task score. The joint improvement links higher task rewards to more completed goals in an environment where actions determine subsequent observations. \emph{The benefit of transition-aware training thus extends to sequential decisions grounded in external environment feedback.} Together with TACO, this result establishes TACA's applicability across distinct action spaces and terminal reward functions.

\vspace{-6pt}
\subsection{Training Dynamics}
\label{ssec:training_dynamics}
\vspace{-6pt}

\begin{figure*}[t]
\centering
\includegraphics[width=\textwidth]{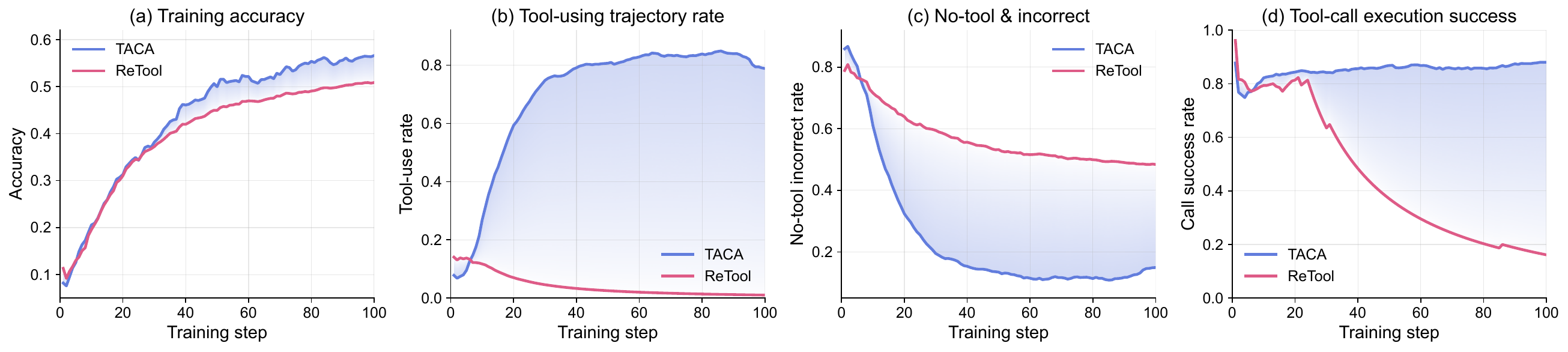}
\vspace{-18pt}
\caption{Training dynamics on Qwen2.5-1.5B-Instruct. (a) Both methods improve training accuracy. TACA maintains (b) substantial tool use, (c) a smaller share of no-tool failures, and (d) a higher logged execution-success metric. Appendix~\ref{app:training_diagnostics} defines the metrics and smoothing.}
\label{fig:sec5_training_dynamics}
\vspace{-16pt}
\end{figure*}

\paragraph{TACA improves training accuracy while retaining tool use.}
Figure~\ref{fig:sec5_training_dynamics}a,b compares learning progress and interaction behavior on Qwen2.5-1.5B-Instruct. Both methods improve training accuracy, but their tool-use rates evolve in opposite directions: TACA maintains substantial tool use, while ReTool increasingly answers without tools. This divergence changes which actions and environment responses enter subsequent training batches. \emph{Improving terminal rewards can accompany the loss of tool interaction, making rollout composition essential to evaluating agentic learning.} TACA sustains an interactive policy while attaining the stronger held-out performance in Table~\ref{tab:main_results_all}.

\vspace{-10pt}
\paragraph{Tool-using trajectories account for most successful solutions.}
Figure~\ref{fig:sec5_training_dynamics}c shows that TACA samples fewer trajectories that both omit tools and answer incorrectly. The joint rollout statistics further connect this shift to successful reasoning: over the displayed training window, tool-using correct trajectories constitute 42.00\% of all TACA trajectories and 87.34\% of its correct trajectories (Appendix~\ref{app:training_diagnostics}). The sustained tool-using mode therefore provides a substantial population of correct interaction trajectories in the training batches. \emph{Tool interaction remains embedded in the model's successful solution strategies as training progresses.} This connects the maintained action support in Figure~\ref{fig:sec5_training_dynamics}b to the task outcomes in panels (a) and (c).

\vspace{-10pt}
\paragraph{Sustained interaction supplies executable tool calls.}
Figure~\ref{fig:sec5_training_dynamics}d tracks the logged execution-success metric alongside the tool-use dynamics. Pooling actual calls over the same training window, TACA executes 47,391 of 54,924 calls successfully, an execution-success rate of 86.28\%, compared with 79.45\% for ReTool (Appendix~\ref{app:training_diagnostics}). The sampled interactions therefore provide a continuing source of executable actions and environment responses. \emph{TACA maintains both the availability of tool interaction and the ability to carry it out, preserving opportunities to learn from external feedback.}

\vspace{-6pt}
\subsection{Ablation Studies}
\label{ssec:ablations}
\vspace{-6pt}

\begin{table}[t]
\centering
\caption{Component ablations on Qwen2.5-1.5B-Instruct. Scores are reported in accuracy (\%).}
\label{tab:ablations}
\begingroup
\setlength{\aboverulesep}{2pt}
\setlength{\belowrulesep}{2pt}
\fontsize{9}{11}\selectfont
\renewcommand{\arraystretch}{1.0}
\setlength{\tabcolsep}{0pt}
\begin{tabular}{@{}p{95pt}>{\centering\arraybackslash}p{46pt}>{\centering\arraybackslash}p{35pt}>{\centering\arraybackslash}p{35pt}>{\centering\arraybackslash}p{34pt}>{\centering\arraybackslash}p{63pt}>{\centering\arraybackslash}p{33pt}>{\centering\arraybackslash}p{25pt}>{\centering\arraybackslash}p{30pt}@{}}
\toprule
\textbf{Variant} & MATH500 & AIME24 & AIME25 & AMC23 & OlympiadBench & GSM8K & Avg. & $\Delta$Avg. \\
\midrule
\rowcolor{TACAtint}
\textbf{TACA (full)} & \textbf{59.14} & \textbf{8.43} & \textbf{8.12} & \textbf{27.42} & \textbf{22.20} & \textbf{74.15} & \textbf{33.24} & -- \\
w/o Advantage Projection & 53.74 & 6.55 & 3.49 & 25.98 & 19.16 & 72.71 & 30.27 & $-2.97$ \\
w/o Decision Contrast & 53.35 & 6.74 & 5.24 & 26.58 & 18.89 & 72.67 & 30.58 & $-2.66$ \\
w/o Supervised Replay & 52.68 & 3.58 & 3.36 & 27.31 & 20.95 & 73.37 & 30.21 & $-3.03$ \\
\bottomrule
\end{tabular}
\endgroup
\vspace{-16pt}
\end{table}

\paragraph{Both advantage adjustments contribute beyond replay supervision.}
Table~\ref{tab:ablations} removes one component at a time from TACA on Qwen2.5-1.5B. Removing Advantage Projection lowers average accuracy from 33.24\% to 30.27\%; removing Decision Contrast lowers it to 30.58\%. Both removals reduce accuracy on every benchmark, including MATH500 drops of 5.40 and 5.79 percentage points. Since Supervised Replay remains active in both variants, these comparisons isolate gains from each advantage adjustment in the presence of replay. \emph{Projecting tool-call advantages and guiding tool-use decisions through reward comparisons each improve performance alongside replay supervision.} This supports treating tool-call credit and response-mode selection as complementary parts of optimization.

\vspace{-10pt}
\paragraph{Supervised Replay adds gains to both advantage adjustments.}
Removing Supervised Replay reduces mean accuracy by 3.03 percentage points and lowers all six benchmark scores (Table~\ref{tab:ablations}). The largest losses occur on MATH500, AIME24, and AIME25, at 6.46, 4.85, and 4.76 points, respectively. With both advantage adjustments enabled, replay adds supervision on stored calls across policy updates. \emph{Replay supervision complements reward-based credit assignment by retaining a learning signal for tool-call generation across changing rollout distributions.} The full method combines this persistent supervision with the prompt-specific adjustments supplied by the current batch.

\begin{table}[t]
\centering
\caption{Mathematical reasoning accuracy (\%) across model size and family. Avg. is the six-benchmark mean; bold marks the best result within each model.}
\label{tab:additional_models}
\begingroup
\setlength{\aboverulesep}{2pt}
\setlength{\belowrulesep}{2pt}
\fontsize{9}{11}\selectfont
\renewcommand{\arraystretch}{1.0}
\setlength{\tabcolsep}{0pt}
\begin{tabular}{@{}p{78pt}>{\centering\arraybackslash}p{46pt}>{\centering\arraybackslash}p{40pt}>{\centering\arraybackslash}p{40pt}>{\centering\arraybackslash}p{40pt}>{\centering\arraybackslash}p{70pt}>{\centering\arraybackslash}p{41pt}>{\centering\arraybackslash}p{41pt}@{}}
\toprule
\textbf{Method} & MATH500 & AIME24 & AIME25 & AMC23 & OlympiadBench & GSM8K & Avg. \\
\midrule
\multicolumn{8}{@{}l}{\textbf{Qwen2.5-7B-Instruct}} \\
MT-GRPO & 68.92 & 11.86 & 8.02 & 44.11 & 33.77 & \textbf{88.84} & 42.59 \\
\rowcolor{TACAtint}
\textbf{TACA} & \textbf{70.77} & \textbf{18.35} & \textbf{13.43} & \textbf{47.78} & \textbf{36.97} & 88.51 & \textbf{45.97} \\
\addlinespace[3pt]\midrule
\multicolumn{8}{@{}l}{\textbf{Llama-3.2-1B-Instruct}} \\
MT-GRPO & 24.04 & 0.79 & 0.00 & 10.12 & 5.14 & 41.58 & 13.61 \\
\rowcolor{TACAtint}
\textbf{TACA} & \textbf{28.07} & \textbf{1.74} & \textbf{0.31} & \textbf{10.93} & \textbf{6.42} & \textbf{43.54} & \textbf{15.17} \\
\bottomrule
\end{tabular}
\endgroup
\vspace{-16pt}
\end{table}

\vspace{-10pt}
\paragraph{The gains persist across model size and family.}
Table~\ref{tab:additional_models} extends the MT-GRPO comparison to Qwen2.5-7B-Instruct and Llama-3.2-1B-Instruct. TACA raises average accuracy from 42.59\% to 45.97\% on the 7B model and from 13.61\% to 15.17\% on Llama, improving five of six and all six benchmarks, respectively. These results complement the 1.5B and 3B comparisons: average gains hold for all four evaluated models. \emph{Transition-aware credit assignment remains useful both as model capacity increases and when the underlying model family changes.}

\section{Related Work}
\label{sec.related}
\vspace{-4pt}
\paragraph{Agentic learning for LLMs.}
Agentic LLMs combine reasoning with tool interaction and iterative feedback, using observations and critiques to guide subsequent generation~\citep{yao2022react,shinn2023reflexion,madaan2023selfrefine,gou2023critic}. Tool-use behavior can be learned through self-supervision, demonstrations, and human feedback~\citep{schick2023toolformer,nakano2021webgpt}, while policy-gradient methods optimize reasoning policies using task rewards~\citep{schulman2017proximal,shao2024deepseekmath,yu2025dapo}. Supervision can assess complete solutions through learned verifiers~\citep{cobbe2021training} or intermediate reasoning steps through process feedback~\citep{uesato2022processoutcome,lightman2023verify}. Recent work extends reinforcement learning to multi-turn tool use and interactive environments, where generated actions determine the observations available for subsequent reasoning~\citep{xue2025simpletir,jin2025searchr1,wang2025ragen,feng2025retool,zhang2025tooln1,shang2025rstar2,mai2025artist,zhang2025agentrl}. Within these systems, reward design and credit assignment govern how task feedback supervises the sequence of actions leading to an outcome~\citep{qian2025toolrl,feng2025gigpo,wang2025practitioner,yu2025demystifying}.

\vspace{-8pt}
\paragraph{Training degradation and collapse in agentic RL.}
Training degradation in agentic RL affects both interaction behavior and policy optimization. Reported patterns include empty turns and gradient spikes~\citep{xue2025simpletir,wang2025practitioner}, repetitive reasoning and reduced exploration~\citep{wang2025ragen,yu2025demystifying}, unstable entropy dynamics~\citep{xu2025epo,dong2025aepo}, and declining response likelihoods~\citep{deng2025lld}. Stabilization methods adapt the granularity of policy updates and control off-policy variance~\citep{li2025sorl,gu2025dspo}, while local credit-assignment methods refine the learning signals received by individual interaction steps~\citep{feng2025gigpo,wei2025turnlevel,wang2025igpo}. Related studies examine how exploration and multi-task training influence the trajectories available for learning~\citep{zhang2025agentrl,jin2025searchr1}. Our analysis connects rollout structure to the aggregate terminal-reward credit assigned to tool-call tokens. We characterize when this credit becomes negative and how trajectory weighting and failure-driven retries affect its allocation during multi-turn optimization. Appendix~\ref{app.related} provides a detailed discussion.

\section{Conclusion}
\label{sec.conclusion}
\vspace{-4pt}
We analyzed training degradation in multi-turn agentic reinforcement learning through the credit assigned to tool interactions. In MT-GRPO, incorrect trajectories containing more tool tokens per turn than correct trajectories can produce negative aggregate tool credit. Token-mean aggregation can amplify this imbalance through length weighting, while failure-driven retries extend terminal supervision to subsequent recovery attempts, including successful tool executions. Motivated by these findings, we proposed Transition-Aware Credit Assignment (TACA), which adjusts advantages for tool-call generation and tool-use decisions and sustains supervision through replay. Experiments across model sizes, families, and tasks demonstrate improved performance and sustained tool interaction, with ablations supporting the complementary contributions of advantage adjustments and replay supervision. These findings suggest that the effectiveness of terminal-reward training depends on how credit is allocated to the intermediate transitions between reasoning and tool interaction.

\newpage
\section*{Acknowledgements}
XYL, ZKZ, JZN, CTC, and BH were supported by NSFC Major Research Plan No. 92570109, RGC Research Fellow Scheme No. RFS2627-2S02, RGC Young Collaborative Research Grant No. C2005-24Y, RGC General Research Fund No. 12202026, RGC General Research Fund No. 12200725, NSFC General Program No. 62376235, Guangdong and Hong Kong Universities “1+1+1” Joint Research Collaboration Scheme, HKBU Faculty Niche Research Areas No. RC-FNRA-IG/25-26/SCI/05, and HKBU CSD Departmental Incentive Scheme. JCY is supported by National Natural Science Foundation of China (No. 62676245), the Lingang Laboratory (Grant No. LGL-8889-01), STCSM (No. 22DZ2229005) and 111 plan (No. BP0719010).

\bibliographystyle{plainnat}
\bibliography{ref}

\onecolumn
\appendix
\part{Appendix}
\etocdepthtag.toc{mtappendix}
\etocsettagdepth{mtchapter}{none}
\etocsettagdepth{mtappendix}{subsection}
\renewcommand{\contentsname}{}
\tableofcontents

\newpage
\section{Ethics Statement}
This study examines reinforcement learning for tool-using language models using existing benchmarks. The experiments use automated task evaluation and involve no human participants.

\section{LLM Usage Disclosure}
\label{app:llm_usage_disclosure}
This manuscript was prepared with the assistance of LLMs, which were used for refining content and ensuring grammatical accuracy. The authors take full responsibility for the entire content of the manuscript.

\section{Impact Statement}
\label{app:impact_statement}
 This work contributes a diagnostic account and mitigation of training collapse in multi-turn agentic reinforcement learning with tool use. By identifying transition-blind credit assignment as a structural cause and proposing Transition-Aware Credit Assignment (TACA), we aim to make agentic post-training of language models more reliable and reproducible, reducing wasted compute on silently collapsing runs and lowering the barrier for smaller research groups developing tool-using reasoning systems.

\section{Limitations}
\label{app:limitations}

The experiments cover mathematical reasoning, TACO coding, and WebShop with Qwen models up to 7B and Llama-3.2-1B. Larger models, long-horizon software engineering, multi-agent coordination, and variability across training seeds are directions for further evaluation. Reward contrasts depend on sufficient sampled support, and Supervised Replay depends on the quality of recorded tool calls. Studying convergence of the full regularized objective is a direction for theoretical development.
\section{Related Work}
\label{app.related}

\paragraph{Reasoning and outcome supervision.}
Chain-of-thought prompting elicits intermediate reasoning steps~\citep{wei2022chainofthought}. Errors in demonstration rationales can reduce answer accuracy~\citep{zhou2024can}, and visualizations of sampled reasoning paths reveal where correct and incorrect solutions diverge~\citep{zhou2026landscape}. At the level of complete solutions, learned verifiers rank candidates by predicted correctness~\citep{cobbe2021training}. PPO optimizes a clipped policy objective~\citep{schulman2017proximal}. DeepSeekMath uses group-relative advantages~\citep{shao2024deepseekmath}, and DAPO aggregates the loss over generated tokens~\citep{yu2025dapo}. In the multi-turn GRPO setting studied here, each tool call inherits the advantage computed from the trajectory's final outcome.

\paragraph{Tool interaction and iterative feedback.}
Interactive reasoning allows a model to obtain missing information by asking a question~\citep{zhou2025from} or acting on an external environment~\citep{yao2022react}. Models learn to browse from demonstrations and human feedback~\citep{nakano2021webgpt} and to insert API calls through self-supervision~\citep{schick2023toolformer}. Models can revise a response using their own critiques or tool feedback~\citep{shinn2023reflexion,madaan2023selfrefine,gou2023critic}, or check a proposed answer for safety before deciding whether to return it~\citep{cao2026resa}. In tool use, an execution error becomes part of the context for the next decision. Section~\ref{ssec:closed_loop} examines how retries change the remaining trajectory and the actions that receive terminal-reward supervision.

\paragraph{Multi-turn tool-integrated reinforcement learning.}
Outcome rewards can train a policy to interleave reasoning with code execution~\citep{li2025torl,feng2025retool,xue2025simpletir,mai2025artist} or search~\citep{jin2025searchr1}. Work on tool learning studies how to reward tool use and train the reasoning that precedes a call~\citep{qian2025toolrl,zhang2025tooln1}. Training systems support these interactions through scalable execution and rollout resampling~\citep{shang2025rstar2} or through multi-environment and asynchronous multi-task training~\citep{xi2025agentgymrl,zhang2025agentrl}. Studies of these policies examine how environment complexity, reward sparsity, and exploration affect training~\citep{wang2025practitioner,yu2025demystifying}. Measuring tool use and resource consumption alongside task outcomes helps distinguish agents that reach similar answers through different interactions~\citep{zhou2026alphadiana}.

\paragraph{Reward normalization and loss aggregation.}
Multi-turn implementations differ in the samples used to compute reward statistics. The released implementations of ToRL, SimpleTIR, Search-R1, and AgentGym-RL normalize rewards over complete trajectories and use token-mean loss aggregation~\citep{li2025torl,xue2025simpletir,jin2025searchr1,xi2025agentgymrl}. LightningRL in Agent Lightning~\citep{luo2025agentlightning} and the GRPO baseline released with GiGPO~\citep{feng2025gigpo} instead replicate each episode return across its turns before computing group statistics; their released GRPO implementations also use token-mean aggregation. This turn-expanded normalization is retained by the released GRPO baselines of RLVMR~\citep{zhang2025rlvmr}, SPEAR~\citep{qin2025spear}, and LaMer~\citep{jiang2025lamer}, and by the RL component of EnvRL-GRPO~\citep{wang2026envrl}. Dr.~GRPO removes response-length and reward-standard-deviation normalization to address optimization biases in GRPO~\citep{liu2025understanding}. CoDaPO weights questions using rollout confidence and estimated difficulty~\citep{zhou2026easy}. Section~\ref{sec.preli} formalizes the turn-level normalization and token-mean objective studied here. Our analysis separates the resulting turn-weighted reward statistics from the token-proportional trajectory weights introduced by loss aggregation, and characterizes their implications for aggregate credit on tool-call spans.

\paragraph{Process supervision and local credit assignment.}
Process supervision evaluates intermediate reasoning steps, including through reward models trained on human step-level feedback~\citep{uesato2022processoutcome,lightman2023verify}. In molecular optimization, supervision can take the form of a reference molecule for a proposed solution~\citep{li2026reference} or a change in molecular properties across successive edits~\citep{li2026learning}. The granularity of evaluation also matters in neural architecture search, where correlations between feature maps can be used to score candidate networks and their constituent operations before training~\citep{jiang2023meco,jiang2025accelerating}. In agentic RL, local credit can be inferred from sampled interactions by grouping actions taken in the same state~\citep{feng2025gigpo} or propagating outcome information through a graph of visited states~\citep{feng2026rewardflow}. Other approaches combine episode-level and local rewards~\citep{zhang2025rlvmr,wang2026selfsum}, vary the timing of feedback~\citep{wei2025turnlevel}, or reward a turn according to the change in the probability of the correct answer~\citep{wang2025igpo}. TACA uses terminal rewards to adjust advantages on tool-call and decision tokens and draws on calls from correct trajectories for replay supervision.

\paragraph{Training degradation and stabilization.}
In multi-turn RL, training degradation can manifest in sampled interactions. RAGEN reports an ``echo trap'' in which reasoning becomes repetitive and trajectory diversity declines~\citep{wang2025ragen}. SimpleTIR links low-probability generations to gradient spikes and filters trajectories containing turns with neither complete code nor a final answer~\citep{xue2025simpletir}. EPO addresses early low-entropy convergence and later instability in exploration through adaptive entropy regularization~\citep{xu2025epo}; AEPO balances entropy in rollout allocation and policy updates~\citep{dong2025aepo}. DSPO combines sequence-level optimization with dynamic sample filtering~\citep{gu2025dspo}, while SORL uses turn-level importance sampling and clipping-triggered normalization to stabilize off-policy updates~\citep{li2025sorl}. \citet{deng2025lld} identify stagnating or declining likelihoods of both correct and incorrect responses and introduce likelihood-preserving regularization. \citet{hao2026toolcollapse} associate failures of tool-invocation structure with probability spikes in control tokens and study supervisory signals for stabilizing training.

Our analysis concerns the aggregate terminal-reward credit assigned to tool-call tokens. For mixed-outcome groups with binary rewards, the fully on-policy objective in Section~\ref{sec.preli} assigns negative aggregate tool credit when incorrect trajectories contain more tool tokens per generated turn than correct trajectories. We further characterize the credit shift induced by token-mean aggregation and examine the additional actions contributed by failure-conditioned retries. These relationships motivate TACA's three components: Advantage Projection, Decision Contrast, and Supervised Replay. The outer objective retains token-proportional trajectory weighting; the corrections act on the credit assigned within that objective.

\section{Extended Analysis}
\label{app:extended_analysis}

We derive the credit identities and length-weighting relationships in Section~\ref{sec.analysis}. We use the reward statistics and generated-token objective of Section~\ref{sec.preli}. Unless a batch is specified, quantities refer to one prompt group. Advantages are held fixed when differentiating the policy surrogate, and standard deviations use the population convention in Eq.~\eqref{eq:turn_reward_statistics}.

\subsection{Terminal-Reward Coefficients in the Policy Gradient}
\label{app:detailed_four_cell}

Let $y_{i,h,t}$ be the generated token at position $t$ in turn $h$ of trajectory $i$, and let $H_{i,h,t}$ contain its prompt and preceding interaction history. At the rollout policy, the token importance ratio $\rho_{i,h,t}$ equals one and lies inside the clipping interval. For the clipped surrogate $\phi$ with turn-normalized terminal-reward advantage $A_i^{\mathrm{turn}}$, the derivative is
\begin{equation}
\left.\nabla_\theta\phi(\rho_{i,h,t},A_i^{\mathrm{turn}})
\right|_{\theta=\theta_{\mathrm{old}}}
=A_i^{\mathrm{turn}}
\left.\nabla_\theta\log\pi_\theta(y_{i,h,t}\mid H_{i,h,t})
\right|_{\theta=\theta_{\mathrm{old}}}.
\label{eq:reward_surrogate_gradient}
\end{equation}
For a subset $\mathcal Z$ of policy-scored generated tokens, its contribution to the on-policy reward-surrogate gradient, normalized by the group's total policy-scored token count $N$, is
\begin{equation}
g_{\mathcal Z}
=\frac1N\sum_{(i,h,t)\in\mathcal Z}A_i^{\mathrm{turn}}
\left.\nabla_\theta\log\pi_\theta(y_{i,h,t}\mid H_{i,h,t})
\right|_{\theta=\theta_{\mathrm{old}}}.
\label{eq:tool_gradient_main}
\end{equation}
This gives the coefficient $A_i^{\mathrm{turn}}/N$ used in Proposition~\ref{prop:tool_credit}. Within a prompt group, the mean and standard deviation are shared, so equal terminal rewards produce equal advantage coefficients. In a mixed binary-reward group, every tool call in an incorrect trajectory receives a negative coefficient, including calls whose execution succeeds. Execution feedback enters subsequent generation contexts; the completed trajectory's reward determines the coefficient assigned to each generated token.

\subsection{Proof of the Tool-Credit Condition}
\label{app:detailed_token_mean}

For a subset $\mathcal Z$ of policy-scored generated tokens, let $z_{i,h}$ count its tokens in turn $h$ and let $z_i=\sum_hz_{i,h}$. Summing the coefficients in Eq.~\eqref{eq:tool_gradient_main} defines its aggregate credit:
\begin{equation}
C(\mathcal Z)
\coloneqq\frac1N\sum_i z_iA_i^{\mathrm{turn}}.
\label{eq:subset_credit_def}
\end{equation}
With $S_i$ generated turns in trajectory $i$ and $M=\sum_i S_i$ turns in the group, $\bar\ell=N/M$ is the mean generated-token count per turn. For a mixed group with binary terminal rewards $R_i$, define the correct and incorrect trajectory sets $\mathcal P=\{i:R_i=1\}$ and $\mathcal F=\{i:R_i=0\}$, their turn counts $M^+=\sum_{i\in\mathcal P}S_i$ and $M^-=\sum_{i\in\mathcal F}S_i$, and their subset-token densities
\[
\bar z^+_{\mathrm{turn}}=\frac{\sum_{i\in\mathcal P}z_i}{M^+},
\qquad
\bar z^-_{\mathrm{turn}}=\frac{\sum_{i\in\mathcal F}z_i}{M^-}.
\]

\begin{lemma}[Subset-credit factorization]
\label{thm:subset_credit}
For a prompt group containing both binary outcomes, with turn-weighted reward standard deviation $\sigma_{\mathrm{turn}}$,
\begin{equation}
C(\mathcal Z)
=\frac{\sqrt{M^+M^-}}{N}
 (\bar z^+_{\mathrm{turn}}-\bar z^-_{\mathrm{turn}})
=\frac{\sigma_{\mathrm{turn}}}{\bar\ell}
 (\bar z^+_{\mathrm{turn}}-\bar z^-_{\mathrm{turn}}).
\label{eq:subset_bias_closed}
\end{equation}
\end{lemma}
\begin{proof}
The positive mass in the turn-replicated reward distribution is $p=M^+/M$. Thus $\sigma_{\mathrm{turn}}^2=p(1-p)$ and the advantages of correct and incorrect trajectories are, respectively,
\[
A^+_{\mathrm{turn}}=\sqrt{\frac{M^-}{M^+}},
\qquad
A^-_{\mathrm{turn}}=-\sqrt{\frac{M^+}{M^-}}.
\]
Substituting into Eq.~\eqref{eq:subset_credit_def} gives
\begin{align*}
NC(\mathcal Z)
&=M^+\bar z^+_{\mathrm{turn}}\sqrt{\frac{M^-}{M^+}}
 -M^-\bar z^-_{\mathrm{turn}}\sqrt{\frac{M^+}{M^-}}\\
&=\sqrt{M^+M^-}
 (\bar z^+_{\mathrm{turn}}-\bar z^-_{\mathrm{turn}}).
\end{align*}
The second equality follows from $\sigma_{\mathrm{turn}}=\sqrt{M^+M^-}/M$ and $\bar\ell=N/M$.
\end{proof}

Taking $\mathcal Z$ to contain all tool-call tokens gives $z_i=m_i$, the tool-token count in trajectory $i$, and proves Proposition~\ref{prop:tool_credit}. Since $\sigma_{\mathrm{turn}}/\bar\ell>0$, the density difference determines the sign; reward composition and mean turn length scale its magnitude. The factorization also applies to subsets of tool-call tokens, including their opening markers.

\paragraph{Extension to scalar rewards.}
Let $\mathbb E_{\mathrm{turn}}$ and $\operatorname{Cov}_{\mathrm{turn}}$ denote empirical expectation and covariance over the group's $M$ generated turns, each with weight $1/M$. For a bounded scalar terminal reward with $\sigma_{\mathrm{turn}}>0$,
\begin{equation}
\begin{aligned}
C(\mathcal Z)
&=\frac{\mathbb E_{\mathrm{turn}}[z_{i,h}A_i^{\mathrm{turn}}]}{\bar\ell}
=\frac{\operatorname{Cov}_{\mathrm{turn}}(z_{i,h},A_i^{\mathrm{turn}})}{\bar\ell}\\
&=\frac{\operatorname{Cov}_{\mathrm{turn}}(z_{i,h},R_i)}
 {\bar\ell\,\sigma_{\mathrm{turn}}}.
\end{aligned}
\label{eq:scalar_subset_credit}
\end{equation}
The second equality uses $\mathbb E_{\mathrm{turn}}[A_i^{\mathrm{turn}}]=0$, and the third substitutes the standardized reward. Aggregate subset credit is negative precisely when subset-token counts per turn have negative covariance with terminal reward. For binary rewards, this covariance is $p(1-p)(\bar z^+_{\mathrm{turn}}-\bar z^-_{\mathrm{turn}})$, recovering Lemma~\ref{thm:subset_credit}.

\paragraph{Normalization implementations.}
For advantages of the form $A_i=(R_i-\mu_{\mathrm{turn}})/s$ with a shared positive denominator $s$, Eq.~\eqref{eq:scalar_subset_credit} has denominator $\bar\ell s$, and the binary multiplier in Eq.~\eqref{eq:subset_bias_closed} is $\sigma_{\mathrm{turn}}^2/(\bar\ell s)$. Sample-standard-deviation normalization and additive numerical stabilization therefore preserve the sign condition. Groups with identical rewards use zero advantages and contribute zero credit.

\subsection{Credit Aggregation across Prompt Groups}
\label{app:batch_credit}

For prompt $x$, let $\mathcal G_x$ denote the set of sampled trajectory indices. Subscript $x$ identifies quantities computed within this group, including its generated-token count $N_x$, subset credit $C_x(\mathcal Z)$, and turn-level advantages $A_{x,i}^{\mathrm{turn}}$. A batch $\mathcal B$ has $N_{\mathcal B}=\sum_xN_x$ generated tokens. Token-mean aggregation gives
\begin{equation}
C_{\mathcal B}(\mathcal Z)
=\frac1{N_{\mathcal B}}\sum_x\sum_{i\in\mathcal G_x}z_iA_{x,i}^{\mathrm{turn}}
=\sum_x\frac{N_x}{N_{\mathcal B}}C_x(\mathcal Z).
\label{eq:batch_subset_credit}
\end{equation}
For binary rewards, let $\mathcal X_{\mathrm{mix}}$ contain the prompt groups with both outcomes. Applying Lemma~\ref{thm:subset_credit} within each group yields
\begin{equation}
C_{\mathcal B}(\mathcal Z)
=\frac1{N_{\mathcal B}}
\sum_{x\in\mathcal X_{\mathrm{mix}}}
M_x\sigma_{x,\mathrm{turn}}
\left(\bar z^+_{x,\mathrm{turn}}-\bar z^-_{x,\mathrm{turn}}\right).
\label{eq:batch_density_factorization}
\end{equation}
The batch sign is determined by the weighted sum of prompt-level density gaps, with positive weights $M_x\sigma_{x,\mathrm{turn}}$. Figure~\ref{fig:sec3_mechanism}(a,b) summarizes pooled tool-token densities, while Figure~\ref{fig:sec3_mechanism}(c) evaluates batch credit using the advantages assigned within each prompt group.

\subsection{Length-Dependent Reward Normalization}
\label{app:detailed_acc_bias}

For a prompt group of $G$ trajectories with binary terminal rewards $R_i\in\{0,1\}$ and both outcomes present, let $K^+$ and $K^-$ count correct and incorrect trajectories. Trajectory $i$ contains $S_i$ generated turns; their total is $M=\sum_iS_i$, with $M^+$ and $M^-$ contributed by correct and incorrect trajectories. The corresponding mean turn counts are $\bar S^\pm=M^\pm/K^\pm$. Writing $q=K^+/G$ for the correct-trajectory fraction and $\Delta=\bar S^-/\bar S^+$ for the ratio of mean turn counts gives
\begin{equation}
\mu_{\mathrm{turn}}
=\frac{1}{M}\sum_iS_iR_i
=\frac{K^+\bar S^+}{K^+\bar S^++K^-\bar S^-}
=\frac{q}{q+(1-q)\Delta}.
\label{eq:turn_mean_weighting}
\end{equation}
The trajectory-weighted reward mean is $\mu_{\mathrm{traj}}=q$, so
\begin{equation}
\mu_{\mathrm{traj}}-\mu_{\mathrm{turn}}
=\frac{q(1-q)(\Delta-1)}{q+(1-q)\Delta}.
\label{eq:turn_mean_gap}
\end{equation}
When $\Delta>1$, incorrect trajectories constitute $1-q$ of sampled trajectories but contribute the larger share $1-\mu_{\mathrm{turn}}$ of generated turns. The difference between these shares is positive and increases with $\Delta$ at fixed $q$. Applying this counting identity to all trajectories in a batch gives the pooled reward means in Figure~\ref{fig:app_reward_normalization}; within each prompt group, the turn-weighted mean enters advantage normalization through Eq.~\eqref{eq:turn_reward_statistics}.

\begin{lemma}[Length-dependent advantage rescaling]
\label{lem:length_reweighting}
For the same mixed binary-reward group, let $A_{\mathrm{turn}}^\pm$ and $A_{\mathrm{traj}}^\pm$ denote the advantages of correct and incorrect trajectories under turn-level and trajectory-level normalization. Then
\begin{equation}
\frac{A^+_{\mathrm{turn}}}{A^+_{\mathrm{traj}}}=\sqrt\Delta,
\qquad
\frac{|A^-_{\mathrm{turn}}|}{|A^-_{\mathrm{traj}}|}=\Delta^{-1/2}.
\label{eq:adv_rescale}
\end{equation}
\end{lemma}
\begin{proof}
For binary rewards with positive mass $p$, the standardized advantages are $\sqrt{(1-p)/p}$ and $-\sqrt{p/(1-p)}$. Trajectory-level normalization uses $p=q$, while turn-level normalization uses $p=\mu_{\mathrm{turn}}$. Substituting
\[
\frac{1-\mu_{\mathrm{turn}}}{\mu_{\mathrm{turn}}}
=\Delta\frac{1-q}{q}
\]
gives both ratios.
\end{proof}

For $\Delta>1$, turn-level normalization increases positive advantages and reduces negative-advantage magnitudes relative to trajectory-level normalization. The aggregation analysis below uses the resulting $A_i^{\mathrm{turn}}$ in both objectives, isolating the contribution of token weighting.

\subsection{Length Weighting and the Aggregation-Induced Credit Shift}
\label{app:token_aggregation}

For $G$ sampled trajectories, let $S_i$ count the generated turns in trajectory $i$ and let $\ell_{i,h}$ count the model-generated tokens included in the objective from turn $h$. Their trajectory and total token counts are $L_i=\sum_{h=1}^{S_i}\ell_{i,h}>0$ and $N=\sum_{i=1}^G L_i$, respectively. Let $\mathcal J_i(\theta)$ be the per-token average of the surrogate and KL terms in trajectory $i$. Regrouping Eq.~\eqref{eq:turn_tok_objective} gives
\begin{equation}
\mathcal J_{\mathrm{MT\text{-}GRPO}}(\theta)
=\sum_{i=1}^G\frac{L_i}{N}\mathcal J_i(\theta),
\qquad
\mathcal J_i(\theta)
=\frac1{L_i}\sum_{h=1}^{S_i}\sum_{t=1}^{\ell_{i,h}}
\left[\phi(\rho_{i,h,t},A_i^{\mathrm{turn}})-\beta\mathbb D_{\mathrm{KL},i,h,t}\right],
\label{eq:implicit_distribution}
\end{equation}
Here $\phi$ is the clipped surrogate, $\rho_{i,h,t}$ is the token importance ratio, and $\beta$ weights the token-level KL term $\mathbb D_{\mathrm{KL},i,h,t}$ from Eq.~\eqref{eq:turn_tok_objective}. Each trajectory receives weight $L_i/N$. Applying this weight to its tool credit per generated token, $(m_i/L_i)A_i^{\mathrm{turn}}$, gives the token-mean expression in Eq.~\eqref{eq:tool_credit_aggregation}, where $m_i$ counts its tool-call tokens. Replacing $L_i/N$ by $1/G$ gives the trajectory-mean expression.

\paragraph{Weight assigned to incorrect trajectories.}
For a prompt group containing both binary outcomes, let $L^\pm$ be the total generated-token count in the correct and incorrect slices, which contain $K^\pm$ trajectories. Their mean token lengths are $\bar L^\pm=L^\pm/K^\pm$, and the group mean is $\bar L=N/G=q\bar L^++(1-q)\bar L^-$, where $q$ is the correct-trajectory fraction. The incorrect slice receives weight
\begin{equation}
\frac{L^-}{N}
=\frac{(1-q)\bar L^-}{q\bar L^++(1-q)\bar L^-},
\qquad
\frac{L^-}{N}-(1-q)
=\frac{q(1-q)(\bar L^--\bar L^+)}{\bar L}.
\label{eq:failure_token_weight}
\end{equation}
Its objective weight exceeds its sampling frequency exactly when incorrect trajectories have greater mean token length. Let $M^\pm$ count generated turns in the two slices, with mean tokens per turn $\bar\ell^\pm=L^\pm/M^\pm$ and mean turn-count ratio $\Delta=(M^-/K^-)/(M^+/K^+)$. Then $\bar L^-/\bar L^+=\Delta\bar\ell^-/\bar\ell^+$, separating the effects of turn counts and tokens per turn.

\paragraph{Proof of the credit-shift identity.}
For a batch $\mathcal B$, let $z_i$ count tokens from subset $\mathcal Z$ of model-generated tokens in trajectory $i$. Its subset credit per generated token is $c_i=z_iA_i^{\mathrm{turn}}/L_i$, where $A_i^{\mathrm{turn}}$ is computed within its prompt group. Let $\mathbb E_{\mathcal B}$ and $\operatorname{Cov}_{\mathcal B}$ denote the empirical mean and covariance with equal weight $1/|\mathcal B|$ per trajectory, and let $\bar L_{\mathcal B}=\mathbb E_{\mathcal B}[L_i]$. The two aggregations give
\begin{equation}
\begin{aligned}
C_{\mathcal B}^{\mathrm{tok}}(\mathcal Z)
&=\frac{\mathbb E_{\mathcal B}[L_ic_i]}{\bar L_{\mathcal B}},
& C_{\mathcal B}^{\mathrm{traj}}(\mathcal Z)
&=\mathbb E_{\mathcal B}[c_i],\\
C_{\mathcal B}^{\mathrm{tok}}(\mathcal Z)
-C_{\mathcal B}^{\mathrm{traj}}(\mathcal Z)
&=\frac{\operatorname{Cov}_{\mathcal B}(L_i,c_i)}{\bar L_{\mathcal B}}.
\end{aligned}
\label{eq:aggregation_covariance}
\end{equation}
The token-mean expression follows from $\sum_iL_ic_i/\sum_iL_i=\mathbb E_{\mathcal B}[L_ic_i]/\bar L_{\mathcal B}$. Subtracting the trajectory mean gives numerator $\mathbb E_{\mathcal B}[L_ic_i]-\mathbb E_{\mathcal B}[L_i]\mathbb E_{\mathcal B}[c_i]$, which is the empirical covariance. Setting $z_i=m_i$, the tool-call token count, recovers Eq.~\eqref{eq:tool_credit_aggregation} and proves Eq.~\eqref{eq:tool_aggregation_covariance}.

Negative covariance lowers token-mean credit relative to trajectory-mean credit. When $C_{\mathcal B}^{\mathrm{traj}}(\mathcal Z)>0$, the exact condition for a reversal to negative credit is
\[
\operatorname{Cov}_{\mathcal B}(L_i,c_i)
<-\bar L_{\mathcal B}C_{\mathcal B}^{\mathrm{traj}}(\mathcal Z).
\]
When trajectory-mean credit is negative, negative covariance increases its magnitude. Figure~\ref{fig:sec3_length_amplification}(a) compares the two aggregations on the same batches and turn-level advantages.

\subsection{Reward Normalization and Loss Reduction}
\label{app:other_configs}

Reward normalization determines the trajectory advantage; loss reduction determines the weight of each token's surrogate. The following configurations make these choices explicit:
\begin{center}
\small
\renewcommand{\arraystretch}{1.15}
\begin{tabular}{@{}llll@{}}
\toprule
Reward statistics & Loss reduction & Per-token weight & Total trajectory weight\\
\midrule
Trajectory & Trajectory mean & $1/(GL_i)$ & $1/G$\\
Trajectory & Token mean & $1/N$ & $L_i/N$\\
Turn & Turn mean & $1/(M\ell_{i,h})$ & $S_i/M$\\
Turn & Token mean & $1/N$ & $L_i/N$\\
\bottomrule
\end{tabular}
\end{center}
Here turn-mean reduction averages the per-token mean of every generated turn, with $\ell_{i,h}>0$ for each included turn. For a mixed binary-reward group, $\mathcal P$ and $\mathcal F$ index correct and incorrect trajectories. The corresponding trajectory and turn counts are $K^\pm$ and $M^\pm$. Define the average subset counts per trajectory and the average subset-token fractions:
\[
\begin{aligned}
\bar z^+_{\mathrm{traj}}&=\frac1{K^+}\sum_{i\in\mathcal P}z_i,
&\bar z^-_{\mathrm{traj}}&=\frac1{K^-}\sum_{i\in\mathcal F}z_i,\\
\bar f^+_{\mathrm{traj}}&=\frac1{K^+}\sum_{i\in\mathcal P}\frac{z_i}{L_i},
&\bar f^-_{\mathrm{traj}}&=\frac1{K^-}\sum_{i\in\mathcal F}\frac{z_i}{L_i},\\
\bar f^+_{\mathrm{turn}}&=\frac1{M^+}\sum_{i\in\mathcal P}\sum_{h=1}^{S_i}\frac{z_{i,h}}{\ell_{i,h}},
&\bar f^-_{\mathrm{turn}}&=\frac1{M^-}\sum_{i\in\mathcal F}\sum_{h=1}^{S_i}\frac{z_{i,h}}{\ell_{i,h}}.
\end{aligned}
\]
Let $\sigma_{\mathrm{traj}}=\sqrt{q(1-q)}$, $\bar L=N/G$, and $\bar\ell=N/M$. In the credit expressions below, the first superscript specifies the reward-normalization unit: $\mathrm{traj}$ for trajectories or $\mathrm{turn}$ for generated turns. The second specifies the loss reduction: $\mathrm{unit}$ gives equal weight to each such unit's token-mean contribution, while $\mathrm{tok}$ averages all generated-token contributions. Applying the factorization in Lemma~\ref{thm:subset_credit} with the corresponding normalization and reduction gives
\begin{align}
C^{\mathrm{traj,unit}}(\mathcal Z)
&=\sigma_{\mathrm{traj}}
 (\bar f^+_{\mathrm{traj}}-\bar f^-_{\mathrm{traj}}),
\label{eq:traj_unit_cred}\\
C^{\mathrm{traj,tok}}(\mathcal Z)
&=\frac{\sigma_{\mathrm{traj}}}{\bar L}
 (\bar z^+_{\mathrm{traj}}-\bar z^-_{\mathrm{traj}}),
\label{eq:traj_tok_cred}\\
C^{\mathrm{turn,unit}}(\mathcal Z)
&=\sigma_{\mathrm{turn}}
 (\bar f^+_{\mathrm{turn}}-\bar f^-_{\mathrm{turn}}),
\label{eq:turn_unit_cred}\\
C^{\mathrm{turn,tok}}(\mathcal Z)
&=\frac{\sigma_{\mathrm{turn}}}{\bar\ell}
 (\bar z^+_{\mathrm{turn}}-\bar z^-_{\mathrm{turn}}).
\label{eq:turn_tok_cred}
\end{align}
For a unit-mean reduction, the proof uses each trajectory's or turn's subset-token fraction. For token-mean reduction, it uses the subset-token count divided by mean unit length. Consequently, count differences determine the sign under token means, while within-unit fraction differences determine the sign under unit means. Each row uses its specified reward normalization; Eq.~\eqref{eq:aggregation_covariance} compares reductions with the same advantages.

\subsection{Terminal-Reward Credit on Recovery Attempts}
\label{app:detailed_closed_loop}

Section~\ref{ssec:closed_loop} connects execution feedback to subsequent tool actions. Consider tool calls in the generated turn immediately following an execution failure, corresponding to the continuation event in Figure~\ref{fig:sec3_retry_feedback}(a). In completed trajectory $i$, let $m_i$ count all tool-call tokens and $r_i$ count tokens in these recovery calls. Partitioning its tool tokens into recovery calls and other calls gives
\begin{equation}
C_{\mathrm{tool}}=C_{\mathrm{rec}}+C_{\mathrm{other}},
\qquad
C_{\mathrm{rec}}=\frac1N\sum_i r_iA_i^{\mathrm{turn}},
\qquad
C_{\mathrm{other}}=\frac1N\sum_i(m_i-r_i)A_i^{\mathrm{turn}}.
\label{eq:recovery_credit_decomposition}
\end{equation}
In a mixed binary-reward group, correct and incorrect trajectories contribute $M^+$ and $M^-$ generated turns. Summing recovery credit over the incorrect trajectories $i\in\mathcal F$ gives
\begin{equation}
C_{\mathrm{rec}}^-
=\frac1N\sum_{i\in\mathcal F}r_iA_i^{\mathrm{turn}}
=-\frac1N\sqrt{\frac{M^+}{M^-}}\sum_{i\in\mathcal F}r_i.
\label{eq:recovery_credit_by_outcome}
\end{equation}
It is strictly negative whenever an incorrect trajectory contains recovery tokens. Each token in a recovery call that executes successfully receives this same negative coefficient. Thus a successful recovery action inherits negative credit when the completed trajectory's answer is incorrect.

Recovery attempts also contribute to the generated-turn and generated-token counts used in reward normalization and objective aggregation. Figure~\ref{fig:sec3_retry_feedback}(a,b) measures subsequent calls and generation length conditional on execution outcomes; panel (c) reports retry rates by final correctness. The credit decomposition specifies how the resulting recovery actions enter terminal-reward supervision.

\section{Additional Analysis Details}
\label{app:analysis}

\subsection{Empirical Setup and Analysis Window}
\label{app:detailed_observation}

Figures~\ref{fig:sec3_overview}--\ref{fig:sec3_length_amplification} analyze Qwen2.5-3B-Instruct trained on DAPO-Math with MT-GRPO and token-mean aggregation. Each batch contains 16 prompts with eight rollouts per prompt, and the maximum interaction length is eight environment steps. Validation accuracy measures final-answer correctness. The evaluation trajectories cover MATH-500, AIME24, AIME25, AMC23, OlympiadBench, MinervaMath, and GSM8K.

The figures display training steps 1--500, with one training batch per step. Validation accuracy peaks at 38.95\% at step 200 and reaches 26.64\% at step 500; the non-agentic reference is 35\%. The peak defines the two reporting intervals, steps 1--200 and 201--500.

\subsection{Overview Metrics and Display Conventions}
\label{app:fig_overview}

\paragraph{Trajectory and execution statistics.}
Trajectory-level metrics count each sampled trajectory once. Correctness is determined by the terminal episode reward, with positive reward indicating a correct trajectory. Tool-use rate is the fraction of trajectories containing at least one tool call; execution success rate is the number of successful executions divided by the number of tool calls. Mean tool-call counts are computed separately for correct and incorrect trajectories. In this environment, each trajectory contains one terminal non-tool turn, so tool-call counts equal generated-turn counts minus one. Over steps 1--200 and 201--500, respectively, the raw mean tool-use rates are 67.77\% and 72.10\%, and execution success rates are 69.41\% and 48.41\%. Correct trajectories average 1.134 and 1.158 tool calls, and incorrect trajectories average 1.131 and 2.026 tool calls.

\paragraph{Time-series display and summaries.}
Validation is plotted at recorded checkpoints. Training curves in Figure~\ref{fig:sec3_overview} use exponential moving average (EMA) smoothing. All 500 training steps enter the displayed curves. The maximum ratio of the smoothed incorrect and correct mean-call curves is 1.90 at step 388. Period means and sign frequencies use raw values, weighting each step equally within this training run.

\subsection{Tool-Token Density and Aggregate Credit}
\label{app:tool_credit_measurements}

\paragraph{Token counts.}
For trajectory $i$, $L_i$ counts valid generated response tokens across its turns, and $m_i$ counts tokens in complete recognized tool-call spans, including delimiters. Token membership is determined by overlap between decoded character ranges and recognized tool-call spans. Both counts use the generated-response mask, with the tool mask additionally selecting the tokens included in $m_i$.

\paragraph{Pooled tool-token density.}
Within each training batch, the density for each correctness slice is its total tool-token count divided by its total generated-turn count. Figure~\ref{fig:sec3_mechanism}(a) plots the raw batch-level correct-minus-incorrect difference. The mean batch-level gap is $-24.35$ tokens per turn over steps 1--200 and $-121.90$ over steps 201--500, a factor of 5.01 in magnitude. It is negative in 468 of 500 batches, including 295 of the 300 batches after step 200.

\paragraph{Batch tool credit.}
Figure~\ref{fig:sec3_mechanism}(c) reports aggregate tool credit
\[
\widehat C_{\mathrm{tool}}=\frac{\sum_i m_iA_i}{\sum_iL_i},
\]
where $A_i$ is the advantage assigned by the training implementation after within-prompt reward normalization. Thus the numerator sums tool-token advantage coefficients and the denominator counts all generated response tokens. Among the 500 batches, 404 have negative credit and 96 have positive credit, giving a negative-credit frequency of 80.8\%. The histogram uses bin width 0.004. The cumulative curve interpolates the empirical cumulative proportions at bin edges, zero, and plot bounds with a shape-preserving monotone cubic interpolation. The point at zero retains the exact negative-credit frequency.

\paragraph{Cross-dataset measurements.}
Table~\ref{tab:dataset_gap} contains the measurements plotted in Figure~\ref{fig:sec3_mechanism}(b). The comparison varies the training dataset with the model, tool environment, training budget, and hyperparameters held fixed. Accuracy is training accuracy, and the gap is correct minus incorrect tool tokens per generated turn.

\begin{table}[htbp]
\centering
\small
\renewcommand{\arraystretch}{1.12}
\caption{Tool-token gap across mathematical training datasets. Accuracy is in \%; the gap is correct minus incorrect tool tokens per turn.}
\label{tab:dataset_gap}
\begin{tabular}{lrr}
\toprule
\textbf{Training dataset} & \textbf{Accuracy (\%)} & \textbf{Tool-token gap} \\
\midrule
MATH & 66.41 & $-151.55$ \\
DAPO-Math & 30.59 & $-184.58$ \\
LIMR & 78.13 & $-138.34$ \\
DeepScaleR & 39.45 & $-127.04$ \\
\bottomrule
\end{tabular}
\end{table}

\subsection{Reward Means and Paired Credit Aggregation}
\label{app:length_measurements}

\begin{figure}[htbp]
\centering
\includegraphics[width=0.62\textwidth]{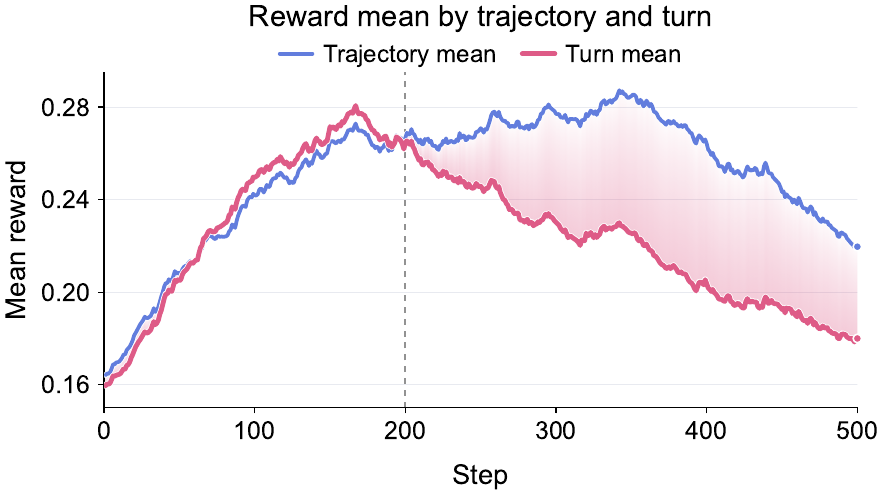}
\caption{Pooled reward means under trajectory and turn weighting. Incorrect trajectories contribute a larger share of generated turns than of sampled trajectories, lowering the turn-weighted mean.}
\label{fig:app_reward_normalization}
\end{figure}

\paragraph{Pooled reward means.}
Figure~\ref{fig:app_reward_normalization} computes reward means over all trajectories in each training batch. The trajectory-weighted mean is the fraction of correct trajectories; the turn-weighted mean is the fraction of generated turns contributed by correct trajectories. Over steps 201--500, their mean values are 0.251745 and 0.197714, respectively, a difference of 5.403 percentage points. The curves use EMA smoothing; interval means are computed from raw batch values.

\paragraph{Paired aggregation.}
Figure~\ref{fig:sec3_length_amplification}(a) applies token-mean and trajectory-mean aggregation to the same batches and their turn-level advantages, using Eq.~\eqref{eq:tool_credit_aggregation}. The curves use EMA smoothing. Over steps 201--500, the raw mean credits are $-0.01704$ and $+0.02323$, respectively. Trajectory-mean credit is positive while token-mean credit is negative in 259 of these 300 batches (86.3\%). This paired calculation evaluates the aggregation-induced credit shift in Eq.~\eqref{eq:tool_aggregation_covariance} with the sampled trajectories and their advantages held fixed.

\paragraph{Paired credit signs.}
Figure~\ref{fig:sec3_length_amplification}(b) groups all 500 batches from steps 1--500 by the signs of their raw aggregate tool credit under the two aggregations. From trajectory mean to token mean, 368 batches (73.6\%) change from positive to negative, and five (1.0\%) change from negative to positive; 91 (18.2\%) remain positive and 36 (7.2\%) remain negative. The negative-credit frequency consequently rises from 8.2\% under trajectory-mean aggregation to 80.8\% under token-mean aggregation. Each connecting band represents the same batches evaluated under both weighting schemes, with its width proportional to their count.

\subsection{Execution-Conditioned Continuation}
\label{app:execution_diagnostics}

\paragraph{Evaluation population.}
Figure~\ref{fig:sec3_retry_feedback}(a,b) uses evaluation trajectories at step 500. The panels pool these trajectories. Calls are classified as successful or failed from tool feedback, with errors and timeouts included among failures. The comparison includes calls with an identified execution outcome.

\paragraph{Next-turn tool-call probability.}
For each execution status and call index, the numerator counts calls followed by a tool call in the next generated turn, and the denominator counts calls with an observed next generated turn. A subsequent tool call is identified by nonempty tool feedback. The call-index groups are 1, 2, 3, and $4+$, with the last group pooling indices 4-7. After the first execution, the next-turn call probabilities are $281/2360=11.9\%$ following success and $2200/2462=89.4\%$ following failure.

\paragraph{Continuation length.}
Figure~\ref{fig:sec3_retry_feedback}(b) counts generated assistant tokens in turns strictly after the first tool-call turn, grouping trajectories by the first execution's status. The groups contain 2,360 trajectories following success and 2,462 following failure. Lengths are obtained by retokenizing the saved assistant response text with the Qwen2.5-3B-Instruct tokenizer, with special-token insertion, padding, and truncation disabled. The median lengths are 107 and 2,048 tokens, and the means are 473.17 and 2,326.64 tokens, respectively. The panel plots empirical cumulative distributions over the full continuation-length range. Its horizontal scale is linear through 500 tokens and logarithmic with base 2 thereafter.

\subsection{Retry Statistics by Final Correctness}
\label{app:retry_statistics}

Table~\ref{tab:retry} gives retry rates and mean generated turns for four mathematics models and two agent environments, grouped by final task correctness. Figure~\ref{fig:sec3_retry_feedback}(c) displays the four MATH/Python retry-rate pairs, with bar heights taken from the table and labels rounded to one decimal place.

\begin{table}[htbp]
\centering
\small
\renewcommand{\arraystretch}{1.12}
\setlength{\tabcolsep}{5pt}
\caption{Reported retry rates and mean rollout turns, separated by terminal correctness. All four named models use MATH/Python; TACO and WebShop use Qwen2.5-1.5B-Instruct in their respective environments.}
\label{tab:retry}
\begin{tabular}{lrrrr}
\toprule
& \multicolumn{2}{c}{\textbf{Retry rate (\%)}} & \multicolumn{2}{c}{\textbf{Mean turns}}\\
\cmidrule(lr){2-3}\cmidrule(lr){4-5}
\textbf{Setting} & Incorrect & Correct & Incorrect & Correct\\
\midrule
\multicolumn{5}{l}{\textit{Mathematical reasoning with Python}} \\
Qwen2.5-1.5B-Instruct & 31.38 & 12.07 & 2.20 & 1.91\\
Qwen2.5-3B-Instruct & 60.55 & 35.96 & 3.15 & 2.38\\
Qwen2.5-7B-Instruct & 36.82 & 21.48 & 1.76 & 1.59\\
Llama-3.2-1B-Instruct & 88.65 & 50.88 & 4.05 & 3.37\\
\midrule
\multicolumn{5}{l}{\textit{Agent tasks with Qwen2.5-1.5B-Instruct}} \\
TACO & 77.03 & 58.68 & 2.96 & 2.44\\
WebShop & 85.34 & 78.14 & 5.29 & 4.81\\
\bottomrule
\end{tabular}
\end{table}

\section{TACA Method Details}
\label{app:taca_method_details}

This section develops the optimization properties of the advantage adjustments in Section~\ref{sec.method} and specifies their integration with replay supervision.

\subsection{Tool-Call Tokens and Decision Positions}
\label{app:taca_token_sets}

For a prompt $x$, let $\mathcal G_x$ index its sampled trajectories, with terminal reward $R_i$ and base GRPO advantage $A_i$ for trajectory $i$. A generated token is indexed by $(i,h,t)$, where $h$ is the generated turn and $t$ the position within that turn. The set $\mathcal Z$ contains the model-generated delimiters and content forming tool calls across $\mathcal G_x$. Preceding reasoning and returned tool observations provide conditioning context; the observations are excluded from the policy objective.

Let $\mathcal T_x\subseteq\mathcal G_x$ contain trajectories invoking at least one tool and $\mathcal N_x=\mathcal G_x\setminus\mathcal T_x$ those answering without tools. The set $\mathcal Z_{\mathrm{dec}}\subseteq\mathcal Z$ contains tokens encoding tool-call opening markers, such as \texttt{<python\_code>}. The set $\mathcal Z_{\mathrm{dir}}$, disjoint from $\mathcal Z$, contains tokens encoding answer-opening markers, such as \texttt{<answer>}, from trajectories in $\mathcal N_x$. With the common tool-token offset $b_{x,\mathcal Z}$, reward contrast $U_x$, and contrast weight $\lambda_{\mathrm{dc}}\geq0$, Eq.~\eqref{eq:taca_advantage} becomes
\begin{equation}
A^{\mathrm{TACA}}_{i,h,t}=
\begin{cases}
A_i+b_{x,\mathcal Z}+\lambda_{\mathrm{dc}}U_x,
 & (i,h,t)\in\mathcal Z_{\mathrm{dec}},\\
A_i+b_{x,\mathcal Z},
 & (i,h,t)\in\mathcal Z\setminus\mathcal Z_{\mathrm{dec}},\\
A_i-\lambda_{\mathrm{dc}}U_x,
 & (i,h,t)\in\mathcal Z_{\mathrm{dir}},\\
A_i, & \text{all other generated tokens}.
\end{cases}
\label{eq:taca_role_advantages}
\end{equation}
Tool-call openings receive both adjustments, while the remaining tool-call tokens receive the same uniform offset. Each trajectory's base advantage remains present in every case.

\subsection{Advantage Projection as a Minimal Adjustment}
\label{app:taca_realized_lift}

For a nonempty tool-call token set $\mathcal Z$, collect the base advantages into a vector $\boldsymbol a\in\mathbb R^{|\mathcal Z|}$, with one entry $A_i$ for each token $(i,h,t)\in\mathcal Z$. Its mean is $\bar A_{x,\mathcal Z}=|\mathcal Z|^{-1}\boldsymbol 1^\top\boldsymbol a$, where $\boldsymbol 1$ is the all-ones vector. Let $\boldsymbol v$ denote a candidate tool-token advantage vector, permitting a distinct adjustment at every position.

\begin{proposition}[Minimum-distance tool-credit adjustment]
\label{prop:taca_projection}
The unique solution of
\begin{equation}
\min_{\boldsymbol v\in\mathbb R^{|\mathcal Z|}}
\frac12\|\boldsymbol v-\boldsymbol a\|_2^2
\quad\text{subject to}\quad
\boldsymbol 1^\top\boldsymbol v\geq0
\label{eq:taca_projection_problem}
\end{equation}
is $\boldsymbol a^{\mathrm{proj}}=\boldsymbol a+b_{x,\mathcal Z}\boldsymbol 1$, with $b_{x,\mathcal Z}=\max\{-\bar A_{x,\mathcal Z},0\}$ as in Eq.~\eqref{eq:zmean_advantage}.
\end{proposition}

\begin{proof}
The objective is strictly convex and the constraint is a halfspace. Introducing the nonnegative Lagrange multiplier $\nu$ for $-\boldsymbol 1^\top\boldsymbol v\leq0$ gives the stationarity condition $\boldsymbol v=\boldsymbol a+\nu\boldsymbol 1$. If $\bar A_{x,\mathcal Z}\geq0$, the original vector is feasible and $\nu=0$. Otherwise, complementary slackness requires $\boldsymbol 1^\top\boldsymbol v=0$, yielding $\nu=-\bar A_{x,\mathcal Z}$. Both cases give the stated solution, and strict convexity establishes uniqueness.
\end{proof}

The projection changes only the uniform component of the advantage vector:
\begin{equation}
\boldsymbol a^{\mathrm{proj}}
-\max\{\bar A_{x,\mathcal Z},0\}\boldsymbol 1
=\boldsymbol a-\bar A_{x,\mathcal Z}\boldsymbol 1.
\label{eq:taca_centered_advantages}
\end{equation}
Consequently, the centered variation and all pairwise advantage differences within the tool-call set are preserved. The squared adjustment is $|\mathcal Z|b_{x,\mathcal Z}^2$. Advantage Projection therefore satisfies the aggregate-credit constraint with the smallest total squared change to the sampled token advantages.

For the MT-GRPO advantages of Section~\ref{sec.preli}, the correction also has a direct expression in terms of the imbalance analyzed in Section~\ref{ssec:token_mean_bias}. For a nonempty tool-call set in a prompt group containing both binary reward outcomes, let $M$ be the total generated-turn count, $\sigma_{\mathrm{turn}}$ the turn-weighted reward standard deviation, and $\bar m_{\mathrm{turn}}^+$ and $\bar m_{\mathrm{turn}}^-$ the tool-token counts per generated turn in correct and incorrect trajectories. Proposition~\ref{prop:tool_credit} gives
\begin{equation}
b_{x,\mathcal Z}
=\frac{M\sigma_{\mathrm{turn}}}{|\mathcal Z|}
\max\{\bar m_{\mathrm{turn}}^- -\bar m_{\mathrm{turn}}^+,0\}.
\label{eq:taca_lift_density}
\end{equation}
Thus the projection offset responds to the outcome-dependent tool-token imbalance through the same quantities that determine aggregate tool credit. Its general definition in Eq.~\eqref{eq:zmean_advantage} applies directly to scalar-reward advantages.

\subsection{Shrinkage and Credit in Decision Contrast}
\label{app:taca_gic_details}

When the tool-using and direct-answer trajectory sets $\mathcal T_x$ and $\mathcal N_x$ are both nonempty, define the empirical reward gap
\[
d_x=\frac{1}{|\mathcal T_x|}\sum_{i\in\mathcal T_x}R_i
-\frac{1}{|\mathcal N_x|}\sum_{i\in\mathcal N_x}R_i.
\]
Each trajectory contributes once to its mode's mean. With $n_x=\min\{|\mathcal T_x|,|\mathcal N_x|\}>0$ and shrinkage parameter $\eta\geq0$, the contrast in Eq.~\eqref{eq:gic} minimizes a regularized quadratic over a candidate scalar contrast $u$:
\begin{equation}
U_x=\underset{u\in\mathbb R}{\operatorname{argmin}}
\left\{\frac{n_x}{2}(u-d_x)^2+\frac{\eta}{2}u^2\right\}
=\frac{n_x}{n_x+\eta}d_x.
\label{eq:taca_contrast_shrinkage}
\end{equation}
Differentiating the quadratic gives $(n_x+\eta)u-n_xd_x=0$. The resulting contrast retains the sign of the empirical reward gap and approaches that gap as both groups grow for fixed $\eta$. When either mode is absent, $U_x=0$.

The smaller group size determines the largest influence of an individual reward on this comparison. For fixed mode memberships,
\begin{equation}
\frac{\partial U_x}{\partial R_i}
=\frac{n_x}{n_x+\eta}
\begin{cases}
|\mathcal T_x|^{-1}, & i\in\mathcal T_x,\\
-|\mathcal N_x|^{-1}, & i\in\mathcal N_x,
\end{cases}
\qquad
\max_{i\in\mathcal G_x}\left|\frac{\partial U_x}{\partial R_i}\right|
=\frac{1}{n_x+\eta}.
\label{eq:taca_contrast_sensitivity}
\end{equation}
Hence changing one trajectory's reward by $\delta$ changes the contrast by at most $|\delta|/(n_x+\eta)$. This characterizes how shrinkage controls individual-sample influence when one response mode is sparsely represented.

For nonempty decision sets, let $\bar A_{x,\mathrm{dec}}$ and $\bar A_{x,\mathrm{dir}}$ be the means of the base token advantages on $\mathcal Z_{\mathrm{dec}}$ and $\mathcal Z_{\mathrm{dir}}$, respectively, and use the superscript $\mathrm{TACA}$ for their means after both adjustments. Averaging Eq.~\eqref{eq:taca_role_advantages} yields
\begin{equation}
\bar A^{\mathrm{TACA}}_{x,\mathrm{dec}}
-\bar A^{\mathrm{TACA}}_{x,\mathrm{dir}}
=\bar A_{x,\mathrm{dec}}-\bar A_{x,\mathrm{dir}}
+b_{x,\mathcal Z}+2\lambda_{\mathrm{dc}}U_x.
\label{eq:taca_decision_gap}
\end{equation}
The projection offset adjusts the tool side, and the contrast shifts the difference by $2\lambda_{\mathrm{dc}}U_x$ according to the empirical preference between response modes. Summing over all tool-call tokens gives the corresponding aggregate identity:
\begin{equation}
\sum_{(i,h,t)\in\mathcal Z}A^{\mathrm{TACA}}_{i,h,t}
=|\mathcal Z|\max\{\bar A_{x,\mathcal Z},0\}
+\lambda_{\mathrm{dc}}|\mathcal Z_{\mathrm{dec}}|U_x.
\label{eq:taca_combined_credit}
\end{equation}
The two terms separate the aggregate-credit adjustment from the signed reward comparison at tool initiation. Within the same prompt group, tokens in the same case of Eq.~\eqref{eq:taca_role_advantages} receive a common adjustment, preserving the base difference $A_i-A_j$ between trajectories $i$ and $j$.

\subsection{Supervised Replay in the Original Interaction Context}
\label{app:taca_anchor_details}

Supervised Replay stores tool-calling responses from trajectories with correct final outcomes, together with the histories under which they were generated. For each stored pair $(c,y)$, $c$ contains the prompt and interaction history preceding the response, including earlier tool observations, and $y$ is the generated response containing the call. At each policy update, eligible context-response pairs are appended to a buffer of capacity $B$; entries are discarded in arrival order when the capacity is exceeded. A replay minibatch $\mathcal D_{\mathrm{rep}}$ is sampled uniformly from the buffer.

For a response $y$, the mask $m_t^{\mathrm{rep}}(c,y)$ equals one at its tool-call token positions and zero elsewhere. The number of supervised tokens is
\[
M_{\mathrm{rep}}=\sum_{(c,y)\in\mathcal D_{\mathrm{rep}}}
\sum_{t=1}^{|y|}m_t^{\mathrm{rep}}(c,y).
\]
Under the current policy $\pi_\theta$, teacher forcing evaluates each recorded tool-call token $y_t$ conditional on $c$ and the recorded response prefix $y_{<t}$. This prefix includes preceding reasoning in the same response. For $M_{\mathrm{rep}}>0$, differentiating Eq.~\eqref{eq:support_anchor} gives the ascent contribution of the replay term:
\begin{equation}
-\lambda_{\mathrm{rep}}\nabla_\theta\mathcal L_{\mathrm{rep}}
=\frac{\lambda_{\mathrm{rep}}}{M_{\mathrm{rep}}}
\sum_{(c,y)\in\mathcal D_{\mathrm{rep}}}
\sum_{t=1}^{|y|}m_t^{\mathrm{rep}}(c,y)
\nabla_\theta\log\pi_\theta(y_t\mid c,y_{<t}),
\label{eq:taca_anchor_gradient}
\end{equation}
where $\lambda_{\mathrm{rep}}\geq0$ is the replay weight. Each tool-call token in the replay minibatch contributes a log-likelihood gradient with coefficient $\lambda_{\mathrm{rep}}/M_{\mathrm{rep}}$, conditioned on its original interaction context and preceding response tokens. The retained responses from correct trajectories provide tool-generation targets across successive policy updates. An empty replay minibatch contributes zero loss.

\subsection{Training Procedure}
\label{app:taca_training_procedure}

Algorithm~\ref{alg:taca} summarizes one TACA update through the three component operations defined in Section~\ref{sec.method}. The set $\mathcal G_x$ indexes the $G$ trajectories sampled for prompt $x$. Empty tool-call sets and replay minibatches contribute zero offset and zero replay loss, respectively; the reward contrast is zero when either trajectory group is empty.

\begin{algorithm}[t!]
\caption{TACA policy update}
\label{alg:taca}
\small
\renewcommand{\algorithmicrequire}{\textbf{Input:}}
\renewcommand{\algorithmicensure}{\textbf{Output:}}
\begin{algorithmic}[1]
\REQUIRE Policy $\pi_\theta$; prompt batch $\mathcal X$; group size $G$; replay buffer $\mathcal R$; $\eta,\lambda_{\mathrm{dc}},\lambda_{\mathrm{rep}}$.
\STATE $\pi_{\theta_{\mathrm{old}}}\gets\pi_\theta$
\FOR{$x\in\mathcal X$}
  \STATE Sample $\{\tau_i\}_{i\in\mathcal G_x}\sim\pi_{\theta_{\mathrm{old}}}(\cdot\mid x)$, with $|\mathcal G_x|=G$
  \STATE Compute terminal rewards $R_i$ and advantages $A_i=A_i^{\mathrm{turn}}$ (Section~\ref{sec.preli})
  \STATE Form tool-using and direct-answer trajectory sets $\mathcal T_x$ and $\mathcal N_x$
  \STATE Identify tool-call tokens $\mathcal Z$ and decision tokens $\mathcal Z_{\mathrm{dec}},\mathcal Z_{\mathrm{dir}}$
  \STATE \textit{// Advantage Projection: aggregate tool-credit adjustment}
  \STATE $b_{x,\mathcal Z}\gets\max\{-\bar A_{x,\mathcal Z},0\}$ if $\mathcal Z\ne\varnothing$, else $0$ \hfill Eq.~\eqref{eq:zmean_advantage}
  \STATE \textit{// Decision Contrast: tool-use versus direct-answer rewards}
  \STATE $U_x\gets\operatorname{Decision\,Contrast}(\mathcal T_x,\mathcal N_x,\{R_i\}_{i\in\mathcal G_x};\eta)$ \hfill Eq.~\eqref{eq:gic}
  \STATE $A^{\mathrm{TACA}}_{i,h,t}\gets A_i+b_{x,\mathcal Z}\mathbf1\{(i,h,t)\in\mathcal Z\}+\lambda_{\mathrm{dc}}U_x\left[\mathbf1\{(i,h,t)\in\mathcal Z_{\mathrm{dec}}\}-\mathbf1\{(i,h,t)\in\mathcal Z_{\mathrm{dir}}\}\right]$
\ENDFOR
\STATE \textit{// Supervised Replay: replay supervision for tool-call generation}
\STATE Update $\mathcal R$ with context-response pairs $(c,y)$ from tool-calling turns in sampled trajectories with correct final outcomes
\STATE Uniformly sample a replay minibatch $\mathcal D_{\mathrm{rep}}$ from $\mathcal R$
\STATE $\mathcal L_{\mathrm{rep}}(\theta)\gets\operatorname{Supervised\,Replay}(\pi_\theta,\mathcal D_{\mathrm{rep}})$ \hfill Eq.~\eqref{eq:support_anchor}
\STATE \textit{// Policy update with transition-aware advantages and replay supervision}
\STATE Update $\pi_\theta$ by maximizing $\mathcal J_{\mathrm{TACA}}(\pi_\theta)$ with $A^{\mathrm{TACA}}_{i,h,t}$ and $\mathcal L_{\mathrm{rep}}(\theta)$ \hfill Eq.~\eqref{eq:taca_objective}
\ENSURE Updated policy $\pi_\theta$ and replay buffer $\mathcal R$.
\end{algorithmic}
\end{algorithm}

\begingroup
\raggedbottom
\section{Implementation Details}
\label{app:implementation}

\subsection{Training Configuration}
\label{app:training_configuration}

\paragraph{Models and interaction environment.}
We post-train Qwen2.5-1.5B-Instruct, Qwen2.5-3B-Instruct, Qwen2.5-7B-Instruct~\citep{yang2024qwen2}, and Llama-3.2-1B-Instruct on MATH~\citep{hendrycksmath2021}. Multi-turn training uses AlphaApollo~\citep{zhou2025alphaapollo} on four A800 GPUs, with Python code execution as the primary reasoning tool. Each rollout alternates generated responses with tool observations until a final answer or the six-turn limit. Tool observations provide context; policy updates score model-generated tokens.

\paragraph{Sampling and optimization.}
Each training batch contains 32 prompts and eight trajectories per prompt, yielding 256 trajectories. Rollouts use temperature 1.0 and top-p 1.0. We optimize with AdamW at a constant learning rate of $10^{-6}$, with one policy-optimization epoch per rollout batch. The implementation applies dual clipping with coefficient 3.0 to negative-advantage updates. Table~\ref{tab:training_configuration} gives the sampling budgets and optimization parameters. The response budget is 1,024 tokens per turn for Qwen2.5-7B-Instruct and 2,048 for the other backbones. The policy-training seed is 42.

\begin{table}[htbp]
\centering
\caption{Configuration for the mathematical reasoning experiments. TACA coefficients are shared across the evaluated backbones. Replay sizes count stored context-response pairs.}
\label{tab:training_configuration}
\begingroup
\fontsize{9}{11}\selectfont
\renewcommand{\arraystretch}{1.08}
\setlength{\tabcolsep}{0pt}
\begin{tabular}{@{}>{\raggedright\arraybackslash}m{210pt}>{\raggedright\arraybackslash}m{186pt}@{}}
\toprule
\textbf{Parameter} & \textbf{Setting} \\
\midrule
\multicolumn{2}{@{}l}{\textit{Rollout configuration}} \\
Prompts per batch / trajectories per prompt & 32 / 8 \\
Training temperature / top-p & 1.0 / 1.0 \\
Maximum input-context length & 7,168 tokens \\
Maximum generated turns & 6 \\
Response tokens per turn & 1,024 (Qwen2.5-7B); 2,048 (other models) \\
Python execution timeout & 3 seconds \\
\midrule
\multicolumn{2}{@{}l}{\textit{Policy optimization}} \\
Optimizer / learning rate & AdamW / $10^{-6}$ \\
Learning-rate schedule / warm-up & Constant / 0 steps \\
Adam coefficients / weight decay & $(0.9,0.999)$ / 0.01 \\
Gradient-norm clipping & 1.0 \\
Policy epochs per rollout batch & 1 \\
Importance-ratio clipping threshold $\varepsilon$ & 0.2 \\
Negative-advantage dual-clipping parameter & 3.0 \\
KL weight $\beta$ / entropy coefficient & 0 / 0 \\
Policy-training seed & 42 \\
\bottomrule
\end{tabular}
\endgroup
\end{table}

\paragraph{TACA configuration and ablations.}
The replay buffer stores tool-calling responses from trajectories with correct final outcomes, together with their preceding interaction contexts, and supplies up to 32 pairs per policy update, sampled uniformly. Table~\ref{tab:training_configuration} specifies the fixed weights for the reward contrast and replay loss, together with the contrast shrinkage and replay capacity. The ablations in Table~\ref{tab:ablations} disable the Advantage Projection offset, Decision Contrast term, or Supervised Replay loss individually.

\subsection{Tasks, Rewards, and Evaluation}
\label{app:task_evaluation}

\paragraph{Mathematical reasoning benchmarks.}
The evaluation suite comprises six benchmarks: MATH500~\citep{hendrycksmath2021}, a 500-problem subset of the MATH test set; AIME 2024\footnote{\url{https://huggingface.co/datasets/HuggingFaceH4/aime_2024}} and AIME 2025\footnote{\url{https://huggingface.co/datasets/opencompass/AIME2025}}, containing short-answer competition problems; AMC 2023\footnote{\url{https://huggingface.co/datasets/math-ai/amc23}}, containing problems adapted from AMC 12A and AMC 12B in numerical-answer form; OlympiadBench~\citep{he2024olympiadbench}, covering olympiad-level mathematical reasoning; and GSM8K~\citep{cobbe2021training}, covering multi-step arithmetic word problems.

\paragraph{Answer verification and evaluation sampling.}
We use terminal answer correctness as the mathematical reasoning reward~\citep{zeng2025simplerl}. For trajectory $i$, let $a_i$ denote its extracted final answer and $a^{\star}$ the reference answer. Math-Verify\footnote{\url{https://github.com/huggingface/Math-Verify}} checks their mathematical equivalence, assigning terminal reward $R_i=1$ for an equivalent answer and $R_i=0$ otherwise. Evaluation samples 32 responses per question at temperature 0.6 and top-p 1.0. Benchmark scores report mean correctness over sampled responses as a percentage; the overall score is their unweighted six-benchmark mean.

\paragraph{Coding.}
We train Qwen2.5-1.5B-Instruct on TACO~\citep{li2023taco} and evaluate on its test set. The coding environment provides a public-test runner and a scratch Python executor, with unit-test outcomes determining terminal rewards. Table~\ref{tab:task_results} reports unit-test scores separately for HackerRank, HackerEarth, CodeForces, and CodeChef; the TACO average weights the four sources equally.

\paragraph{Interactive decision-making.}
We train Qwen2.5-1.5B-Instruct separately on WebShop~\citep{yao2022webshop} using its official data split. The policy interacts through search, navigation, and purchase actions and receives a terminal task reward. Evaluation reports success rate for completing the shopping goal and the task score supplied by the environment, as shown in Table~\ref{tab:task_results}.

\paragraph{Baseline comparisons.}
Base-model inference and agentic prompting evaluate the starting models without policy updates. The trained baselines comprise non-agentic GRPO~\citep{shao2024deepseekmath}, ReTool~\citep{feng2025retool}, MT-GRPO~\citep{feng2025gigpo}, GiGPO~\citep{feng2025gigpo}, and SimpleTIR~\citep{xue2025simpletir}. The mathematical reasoning comparison uses the six-benchmark protocol above; the coding and interactive comparisons use the corresponding TACO and WebShop metrics. Training and evaluation settings are matched within each model-task pair.

\subsection{Interaction Prompts}
\label{app:interaction_prompts}

The agentic mathematical reasoning experiments use the templates in Figures~\ref{fig:prompt-no-his} and~\ref{fig:prompt-with-his} for training and evaluation. The first-turn template presents the question and action format; later turns additionally receive the generated-turn count and interaction history. Responses place reasoning inside \texttt{<think>} tags and select either a Python call delimited by \texttt{<python\_code>} or a final answer delimited by \texttt{<answer>}. Execution results are returned as tool observations for the next turn.

\begin{figure}[H]
\centering
\begin{tcolorbox}[title = {Prompt Template (No History)}, fontupper=\normalsize\ttfamily]
\footnotesize
\begin{verbatim}
You are a problem solver agent that combines step-by-step reasoning with Python code 
execution to solve problems accurately.

Your question: {question}

You MUST first think inside <think> </think> tags, then you MUST choose exactly ONE of 
the following actions:
1) [RECOMMENDED] <python_code>...</python_code>: Write exactly  ONE block of pure 
    Python 3 to compute, verify, or explore the problem. You SHOULD use this tool 
    to check your reasoning before giving a final answer. Inspect the <tool_response> 
    (stdout from your code). If it disagrees with your reasoning, correct yourself.
2) <answer>...</answer>: Provide your final answer ONLY after you have verified it 
    (ideally via code). You MUST format the answer in LaTeX inside \boxed{...}.

IMPORTANT RULES:
- You MUST use <python_code> at least once to verify your reasoning before providing 
your final <answer>. Skipping code verification leads to errors.
- You MUST NOT output both <python_code> and <answer> in the same step.
- Your <think> block MUST come before any action tag.
\end{verbatim}
\end{tcolorbox}
\caption{Prompt template used for the first step (no interaction history).}
\label{fig:prompt-no-his}
\end{figure}

\begin{figure}[H]
\centering
\begin{tcolorbox}[title = {Prompt Template (With History)}, fontupper=\normalsize\ttfamily]
\footnotesize
\begin{verbatim}
You are a problem solver agent that combines step-by-step reasoning with Python code
execution to solve problems accurately.

Your question: {question}

Prior to this step, you have already taken {step_count} step(s).
Below is the interaction history:
{memory_context}

1) [RECOMMENDED] <python_code>...</python_code>: Write exactly  ONE block of pure 
    Python 3 to compute, verify, or explore the problem. You SHOULD use this tool 
    to check your reasoning before giving a final answer. Inspect the <tool_response> 
    (stdout from your code). If it disagrees with your reasoning, correct yourself.
2) <answer>...</answer>: Provide your final answer ONLY after you have verified it 
    (ideally via code). You MUST format the answer in LaTeX inside \boxed{...}.

IMPORTANT RULES:
- You MUST use <python_code> at least once to verify your reasoning before providing 
your final <answer>. Skipping code verification leads to errors.
- You MUST NOT output both <python_code> and <answer> in the same step.
- Your <think> block MUST come before any action tag.
\end{verbatim}
\end{tcolorbox}
\caption{Prompt template used for subsequent steps (with interaction history).}
\label{fig:prompt-with-his}
\end{figure}

\subsection{Normalization Comparison}
\label{app:normalization_comparison}

\paragraph{Comparison protocol.}
We compare MT-GRPO, Dr.~GRPO~\citep{liu2025understanding}, and TACA on Qwen2.5-1.5B-Instruct using the same MATH training data, Python environment, and six-benchmark evaluation protocol. Dr.~GRPO uses a constant response normalizer and omits reward-standard-deviation normalization. TACA uses the adjusted advantages and replay loss in Eq.~\eqref{eq:taca_objective}.

\begin{table}[t]
\centering
\caption{Comparison with Dr.~GRPO on Qwen2.5-1.5B-Instruct (accuracy, in \%).}
\label{tab:dr_grpo}\label{tab:normalization_comparison}
\begingroup
\fontsize{9}{11}\selectfont
\renewcommand{\arraystretch}{1.08}
\setlength{\tabcolsep}{0pt}
\begin{tabular}{@{}p{96pt}>{\centering\arraybackslash}p{44pt}>{\centering\arraybackslash}p{41pt}>{\centering\arraybackslash}p{41pt}>{\centering\arraybackslash}p{41pt}>{\centering\arraybackslash}p{51pt}>{\centering\arraybackslash}p{41pt}>{\centering\arraybackslash}p{41pt}@{}}
\toprule
\textbf{Method} & MATH500 & AIME24 & AIME25 & AMC23 & \makecell[cc]{Olympiad\\Bench} & GSM8K & Avg. \\
\midrule
MT-GRPO & 50.70 & 3.26 & 0.97 & \textbf{27.50} & 17.88 & 71.49 & 28.63 \\
Dr.~GRPO & 51.76 & 4.03 & 0.89 & 27.40 & 17.57 & 72.56 & 29.04 \\
\rowcolor{TACAtint}
\textbf{TACA} & \textbf{59.14} & \textbf{8.43} & \textbf{8.12} & 27.42 & \textbf{22.20} & \textbf{74.15} & \textbf{33.24} \\
\bottomrule
\end{tabular}
\endgroup
\end{table}

\paragraph{Results.}
Table~\ref{tab:normalization_comparison} gives mean accuracies of 28.63\%, 29.04\%, and 33.24\% for MT-GRPO, Dr.~GRPO, and TACA. TACA gains 4.20 points over Dr.~GRPO and improves all six benchmarks.

\subsection{Training Diagnostics}
\label{app:training_diagnostics}

\paragraph{Trajectory and execution statistics.}
Table~\ref{tab:training_diagnostics} summarizes unsmoothed records for Qwen2.5-1.5B-Instruct on MATH over steps 1--100, comprising 25,600 trajectories per method. Training accuracy, tool-use rate, and no-tool failure rate are the fractions of trajectories that answer correctly, invoke a tool, and answer incorrectly without tools, respectively. Tool-using correct trajectories are reported both as a fraction of all trajectories and as a fraction of correct trajectories. Execution success divides the total successful calls by the total attempted calls over the same window. For rates over all trajectories, the fixed batch size makes per-step and pooled averages equivalent.

\paragraph{Curve construction.}
Figure~\ref{fig:sec5_training_dynamics} displays training steps 1--100 with exponential moving average (EMA) smoothing. Panels (a--c) show the per-step trajectory rates defined above. Panel (d) shows the per-step execution-success rate, with zero assigned to steps without tool calls. The pooled execution rate in Table~\ref{tab:training_diagnostics} is the ratio of total successful calls to total attempted calls over the displayed window.

\begin{table}[t]
\centering
\caption{Unsmoothed training statistics over steps 1--100 on Qwen2.5-1.5B-Instruct.}
\label{tab:training_diagnostics}
\begingroup
\fontsize{9}{11}\selectfont
\renewcommand{\arraystretch}{1.15}
\begin{tabular}{lrr}
\toprule
Metric & ReTool & TACA \\
\midrule
Training accuracy (\%) & 49.17 & 48.09 \\
Tool-use rate (\%) & 1.49 & 75.76 \\
No-tool failure rate (\%) & 49.75 & 18.15 \\
Tool-using correct trajectories (\%) & 0.41 & 42.00 \\
Tool use among correct trajectories (\%) & 0.84 & 87.34 \\
Steps containing calls & 27 & 100 \\
Attempted calls & 730 & 54,924 \\
Successful calls & 580 & 47,391 \\
Execution success (\%) & 79.45 & 86.28 \\
\bottomrule
\end{tabular}
\endgroup
\end{table}
\clearpage
\endgroup

\clearpage
\section{NeurIPS Paper Checklist}

\begin{enumerate}

\item {\bf Claims}
    \item[] Question: Do the main claims made in the abstract and introduction accurately reflect the paper's contributions and scope?
    \item[] Answer: \answerYes{}
    \item[] Justification: We have summarized the position and key contributions of the paper in the abstract and introduction parts.

\item {\bf Limitations}
    \item[] Question: Does the paper discuss the limitations of the work performed by the authors?
    \item[] Answer: \answerYes{}
    \item[] Justification: We discuss the limitations in Appendix~\ref{app:limitations}.
    

\item {\bf Theory assumptions and proofs}
    \item[] Question: For each theoretical result, does the paper provide the full set of assumptions and a complete (and correct) proof?
    \item[] Answer: \answerYes{}
    \item[] Justification: We provide complete proofs in Appendices~\ref{app:extended_analysis} and~\ref{app:taca_method_details}.

    \item {\bf Experimental result reproducibility}
    \item[] Question: Does the paper fully disclose all the information needed to reproduce the main experimental results of the paper to the extent that it affects the main claims and/or conclusions of the paper (regardless of whether the code and data are provided or not)?
    \item[] Answer: \answerYes{}
    \item[] Justification: The experiment details of implementation are introduced in Section~\ref{sec.exp} and Appendix~\ref{app:implementation}.
    

\item {\bf Open access to data and code}
    \item[] Question: Does the paper provide open access to the data and code, with sufficient instructions to faithfully reproduce the main experimental results, as described in supplemental material?
    \item[] Answer: \answerYes{}
    \item[] Justification: We have open-sourced our code at \url{https://github.com/tmlr-group/TACA}.
    

\item {\bf Experimental setting/details}
    \item[] Question: Does the paper specify all the training and test details (e.g., data splits, hyperparameters, how they were chosen, type of optimizer) necessary to understand the results?
    \item[] Answer: \answerYes{}
    \item[] Justification: The experiment details of implementation are introduced in Section~\ref{sec.exp} and Appendix~\ref{app:implementation}.
        

\item {\bf Experiment statistical significance}
    \item[] Question: Does the paper report error bars suitably and correctly defined or other appropriate information about the statistical significance of the experiments?
    \item[] Answer: \answerYes{}
    \item[] Justification: We conduct evaluations and repeat the inference for each question to obtain reliable results.
    

\item {\bf Experiments compute resources}
    \item[] Question: For each experiment, does the paper provide sufficient information on the computer resources (type of compute workers, memory, time of execution) needed to reproduce the experiments?
    \item[] Answer: \answerYes{}
    \item[] Justification: The computed requirements are introduced in Section~\ref{sec.exp} and Appendix~\ref{app:implementation}.
        
    
\item {\bf Code of ethics}
    \item[] Question: Does the research conducted in the paper conform, in every respect, with the NeurIPS Code of Ethics \url{https://neurips.cc/public/EthicsGuidelines}?
    \item[] Answer: \answerYes{}
    \item[] Justification: 
    We have carefully checked the NeurIPS Code of Ethics and confirmed that our paper obeys it.

\item {\bf Broader impacts}
    \item[] Question: Does the paper discuss both potential positive societal impacts and negative societal impacts of the work performed?
    \item[] Answer: \answerYes{}
    \item[] Justification: 
    The broader impacts are introduced in Appendix~\ref{app:impact_statement}.
    
    
\item {\bf Safeguards}
    \item[] Question: Does the paper describe safeguards that have been put in place for responsible release of data or models that have a high risk for misuse (e.g., pre-trained language models, image generators, or scraped datasets)?
    \item[] Answer: \answerNA{}
    \item[] Justification: This work does not release new models or datasets.
    

\item {\bf Licenses for existing assets}
    \item[] Question: Are the creators or original owners of assets (e.g., code, data, models), used in the paper, properly credited and are the license and terms of use explicitly mentioned and properly respected?
    \item[] Answer: \answerYes{}
    \item[] Justification:
    We have introduced the resources of the models and datasets used in this work.
    

\item {\bf New assets}
    \item[] Question: Are new assets introduced in the paper well documented and is the documentation provided alongside the assets?
    \item[] Answer: \answerNA{}
    \item[] Justification: This work does not introduce new assets.
    

\item {\bf Crowdsourcing and research with human subjects}
    \item[] Question: For crowdsourcing experiments and research with human subjects, does the paper include the full text of instructions given to participants and screenshots, if applicable, as well as details about compensation (if any)? 
    \item[] Answer: \answerNA{}
    \item[] Justification: 
    The paper is not about crowdsourcing experiments or research with human subjects.
    

\item {\bf Institutional review board (IRB) approvals or equivalent for research with human subjects}
    \item[] Question: Does the paper describe potential risks incurred by study participants, whether such risks were disclosed to the subjects, and whether Institutional Review Board (IRB) approvals (or an equivalent approval/review based on the requirements of your country or institution) were obtained?
    \item[] Answer: \answerNA{}
    \item[] Justification: 
    The paper is not about research with human subjects.
    

\item {\bf Declaration of LLM usage}
    \item[] Question: Does the paper describe the usage of LLMs if it is an important, original, or non-standard component of the core methods in this research? Note that if the LLM is used only for writing, editing, or formatting purposes and does \emph{not} impact the core methodology, scientific rigor, or originality of the research, declaration is not required.
    \item[] Answer: \answerNA{}
    \item[] Justification: The core method development in this work does not involve LLMs as any important, original, or non-standard components.
    

\end{enumerate}

\end{document}